\pdfoutput=1

\documentclass[11pt]{article}
\usepackage{authblk}

\usepackage[preprint]{acl}

\usepackage{times}
\usepackage{latexsym}

\usepackage[T1]{fontenc}

\usepackage[utf8]{inputenc}

\usepackage{microtype}

\usepackage{inconsolata}

\usepackage{graphicx}
\usepackage{paralist}
\usepackage{multirow}
\usepackage{colortbl}
\usepackage{bm} 
\usepackage{subfigure}
\usepackage{amsmath}
\usepackage{amsfonts}
\usepackage{booktabs} 
\usepackage{tabularx}
\usepackage{makecell}
\usepackage{adjustbox}
\usepackage{listings}
\usepackage[ruled,vlined]{algorithm2e}
\SetAlFnt{\small}

\title{ARF-RLHF: Adaptive Reward-Following for RLHF through Emotion-Driven Self-Supervision and Trace-Biased Dynamic Optimization}

\author[1,2]{Yuxuan Zhang
\thanks{Corresponding author. Email: \texttt{Avenum\_Z@outlook.com}.}}

\affil[1]{South China Normal University}
\affil[2]{University of Aberdeen}

\begin{document}
\maketitle


\begin{abstract}
Current RLHF methods such as PPO and DPO typically reduce human preferences to binary labels, which are costly to obtain and too coarse to reflect individual variation. We observe that expressions of satisfaction and dissatisfaction follow stable linguistic patterns across users, indicating that more informative supervisory signals can be extracted from free-form feedback. Building on this insight, we introduce Adaptive Reward-Following (ARF), which converts natural feedback into continuous preference trajectories and optimizes them using the novel TraceBias algorithm. Across diverse LLMs and preference domains, ARF consistently outperforms PPO and DPO, improving alignment by up to 7.6\%. Our results demonstrate that continuous reward modeling provides a scalable path toward personalized and theoretically grounded RLHF.
\end{abstract}

\begin{figure*}[tb]
  \centering
  \includegraphics[width=\textwidth]{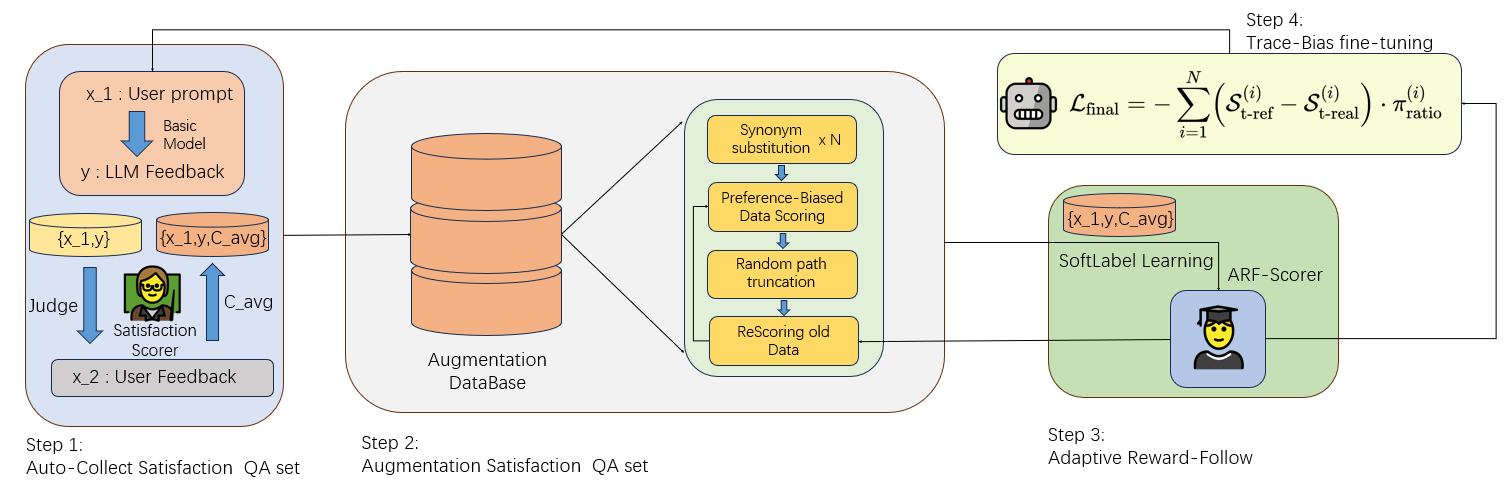}
  \caption{
    Illustrates the overall workflow of our framework. We begin by deriving posterior satisfaction estimates from natural user feedback via a Static Satisfaction Scorer (Step 1). These samples are then stored and augmented through synonym substitution, truncation, and reweighting to form a diversified reward corpus (Step 2). The ARF scorer is trained with soft labels to predict satisfaction scores and is continuously updated (Step 3). Finally, the TraceBias algorithm leverages ARF-generated rewards to fine-tune the LLM (Step 4), completing a fully self-supervised RLHF pipeline.
  }
  \label{fig:Main_Road}
  \vspace{-1em}
\end{figure*}

\section{Introduction}  
The rapid evolution of large language models (LLMs) has shifted the challenge of alignment from factual accuracy to deeper personalization—adapting model behavior to the intent and preferences of individual users. Reinforcement Learning from Human Feedback (RLHF) is now the de facto fine-tuning strategy, powering models such as GPT-4~\cite{openai2024gpt4technicalreport}, DeepSeek-R1~\cite{deepseekai2025deepseekr1incentivizingreasoningcapability}, and Llama-3~\cite{grattafiori2024llama3herdmodels}. However, prevailing RLHF methods such as PPO~\cite{schulman2017proximalpolicyoptimizationalgorithms} and DPO~\cite{rafailov2024directpreferenceoptimizationlanguage} depend on large-scale binary preference annotations. These signals are expensive to collect, coarse in resolution, and biased toward group-level tendencies rather than individual needs.  

Recent efforts to reduce annotation burden—such as RLAIF~\cite{lee2024rlaifvsrlhfscaling}, crowd-sourced preference distillation~\cite{zhang-etal-2024-plad}, and KTO~\cite{ethayarajh2024ktomodelalignmentprospect}—have improved sample efficiency but still rely on externally sourced annotations or handcrafted prompt pipelines. Without rethinking how feedback itself is represented, such approaches cannot capture the evolving trajectory of individual preferences and will continue to inherit population-level biases.  

To address this gap, we draw inspiration from psycholinguistic research. \textit{Common Ground Theory}~\cite{clark1996usinglanguage} shows that people share conventions for signaling satisfaction or dissatisfaction; LIWC analyses~\cite{tausczik2010liwc} confirm stable emotion-word usage across cultures; and \textit{Emotion Prototype Theory}~\cite{fehr1984emotionprototype} explains how affect can be inferred from prototypical linguistic cues. Together, these findings suggest that free-form feedback naturally encodes graded satisfaction signals. We leverage these regularities not as fixed templates, but as priors for interpreting a single user’s feedback in context—transforming open-ended responses into a posterior estimate of satisfaction.  

Guided by this insight, we propose the \textbf{Adaptive Reward-Following (ARF)} framework, a self-supervised RLHF pipeline designed to minimize manual labeling while capturing fine-grained, per-user preferences. As illustrated in Figure~\ref{fig:Main_Road}, ARF:  
\begin{compactenum}
    \item infer posterior satisfaction signals from natural user feedback via a \textit{Static Satisfaction Scorer};  
    \item augment samples through synonym substitution, truncation, and preference-weighted rebalancing;  
    \item train a lightweight \textit{ARF Scorer} with soft labels to directly predict continuous satisfaction scores;  
    \item fine-tune the LLM with the proposed \textit{TraceBias} algorithm, which optimizes reward \emph{trajectories} rather than binary labels.  
\end{compactenum}

This workflow removes the reliance on costly pairwise annotations, while enabling precise modeling of evolving, user-specific preferences. Experiments on Qwen-2/2.5~\cite{qwen2025qwen25technicalreport}, Gemma-2~\cite{gemmateam2024gemma2improvingopen}, and Llama-3.2~\cite{grattafiori2024llama3herdmodels} across four preference domains show that ARF not only matches strong baselines but outperforms PPO by 3.3\% and DPO by 7.6\%.  

\paragraph{Contributions.} Our work makes three key contributions:  
\begin{compactitem}
    \item We introduce ARF, the first self-supervised RLHF framework that learns continuous satisfaction trajectories directly from free-form user feedback.  
    \item We propose \textit{TraceBias}, an RLHF fine-tuning algorithm that leverages these trajectories to optimize alignment without binary comparisons. We also prove its theoretical consistency with standard RLHF methods (see Appendix~\ref{app:Homology}). 
    \item We demonstrate across multiple LLM families and domains that ARF achieves scalable and personalized alignment, reaching or surpassing the performance of several baselines while reducing annotation cost.
\end{compactitem}  

Together, these results position ARF as a step toward next-generation RLHF: alignment that is scalable, personalized, and grounded in both linguistic theory and empirical performance.

\section{Background}
\subsection{RLHF as Preference-Based Optimization}
\label{CoreRLHF}
Reinforcement Learning from Human Feedback (RLHF) has become the dominant paradigm for aligning large language models (LLMs) with human intent. Despite implementation variations, most pipelines share a constant foundation: \textit{(i)} preference modeling via the Bradley–Terry (BT) framework~\cite{10.1093/biomet/42.3-4.450}, and \textit{(ii)} policy optimization using variants of policy gradient.

\paragraph{Pipeline.} 
RLHF typically proceeds in three stages. First, a model is pretrained on large-scale corpora and optionally finetuned (SFT) on instruction data. Second, a preference dataset $\mathcal{D} = \{(x, y_w, y_l)\}$ is collected, where $y_w \succ y_l$ indicates that humans prefer $y_w$ over $y_l$ for input $x$. Third, optimization is performed to maximize expected return under the inferred preference structure.

\paragraph{Preference Modeling.} 
The BT model assumes the probability of preferring $y_w$ to $y_l$ is
\begin{equation}
    \mathcal{P}(y_w \succ y_l | x) = \sigma\!\left(\mathcal{R}(x, y_w) - \mathcal{R}(x, y_l)\right),
\end{equation}
where $\mathcal{R}(x,y)$ is a learned reward function and $\sigma$ the logistic function.

\paragraph{Policy Optimization.}
Given a reward proxy, the policy $\pi_\theta$ is updated using standard policy gradient:
\begin{equation}
    \nabla_\theta J(\theta) = \mathbb{E}_{\tau \sim \pi_\theta} \left[\sum_{t=0}^{T} \nabla_\theta \log \pi_\theta(a_t|s_t)\, A^\pi(s_t,a_t)\right],
\end{equation}
with $A^\pi$ the advantage function. In practice, proximal methods such as PPO~\cite{schulman2017proximalpolicyoptimizationalgorithms} are used to stabilize training.

\subsection{Two Dominant Formulations: PPO and DPO}
Most existing work falls into two closely related approaches:

\paragraph{PPO with Reward Modeling.}
PPO first fits a reward model via BT loss:
\begin{equation}
    \mathcal{L}_{R} = -\mathbb{E}_{(x,y_w,y_l) \sim D} \log \sigma\!\big(\mathcal{R}(x,y_w) - \mathcal{R}(x,y_l)\big),
\end{equation}
and then optimizes the policy with clipped policy gradients.

\paragraph{DPO as a Direct Alternative.}
DPO~\cite{rafailov2024directpreferenceoptimizationlanguage} removes the explicit reward model, directly maximizing the margin between preferred and dispreferred responses:
\begin{align}
    \mathcal{L}_{\text{DPO}}(\theta) &= -\mathbb{E}_{(x,y_w,y_l) \sim D} \log \sigma \Big(
        \beta \log \tfrac{\pi_\theta(y_w|x)}{\pi_{\text{ref}}(y_w|x)} \nonumber \\
        &\quad - \beta \log \tfrac{\pi_\theta(y_l|x)}{\pi_{\text{ref}}(y_l|x)}
    \Big),
\end{align}
where $\pi_{\text{ref}}$ is a reference model and $\beta$ a temperature hyperparameter.

\paragraph{Limitations.}
While PPO/DPO‑based fine‑tuning methods and their variants differ in certain design choices, they are both grounded in the same actor–critic framework (see Appendix~\ref{app:Homology} for an in‑depth discussion) and share common structural constraints: (1) reliance on discrete preference pairs, (2) dependence on costly human annotation, and (3) limited personalization, since alignment is defined at the population level. These limitations motivate approaches that move beyond binary comparisons and static reward proxies.

\begin{figure*}[tb]
    \centering
    \includegraphics[width=\textwidth]{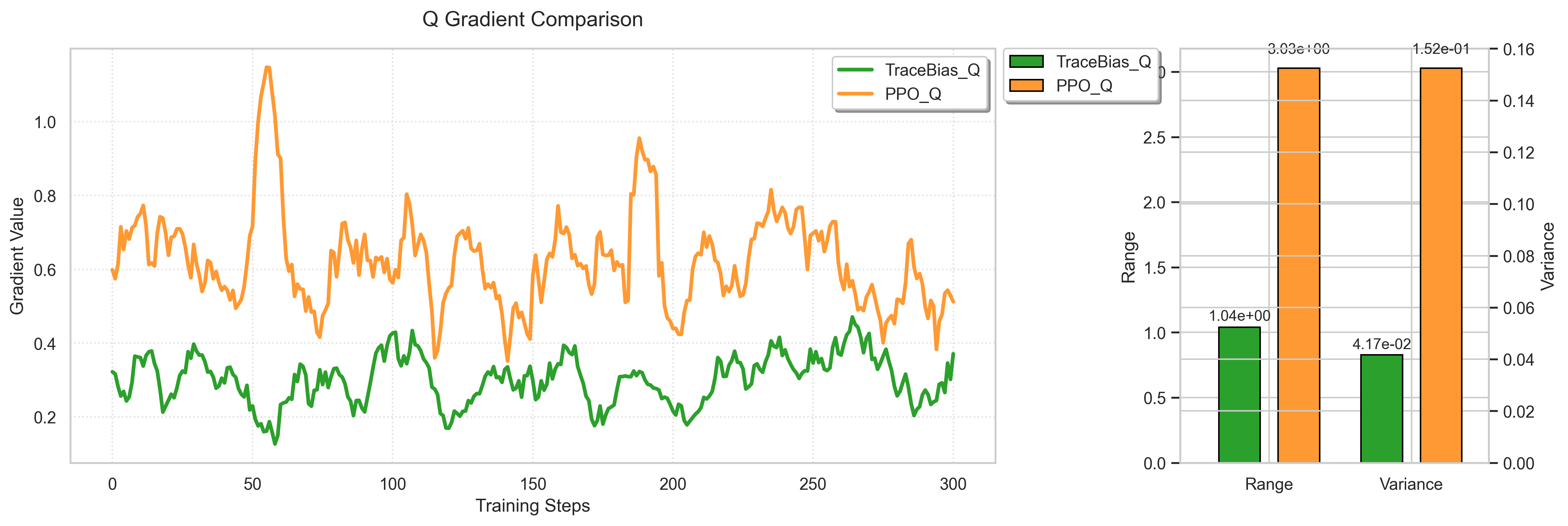}
    \caption{We compare the gradient norm statistics of PPO, using a clip range $\epsilon = 0.2$ as in the original paper~\cite{schulman2017proximalpolicyoptimizationalgorithms} and TraceBias with DAM. DAM exhibits lower variance and more stable gradient magnitudes, suggesting improved training stability and potential for enhanced performance.(V is shown in appendix~\ref{app:V})}
    \label{fig:GradCompare}
    \vspace{-1em}
\end{figure*}

\section{Methodology}
ARF-RLHF builds on the framework in Section~\ref{CoreRLHF} and consists of three stages: model initialization, preference scoring, and policy optimization. For initialization, we use an open-source state-of-the-art LLM. Preference scoring relies on a composite system that combines a high-precision satisfaction scorer with an adaptive ARF scorer that evolves over time. The evolving signals guide policy optimization through the TraceBias algorithm. TraceBias fine-tunes the model with soft reward signals from the scorers, removing the need for binary labels and allowing more nuanced updates.

The following subsections elaborate on each component. We begin with the scoring system, describing the interaction between the static scorer and the adaptive preference tracker. Next, we present the score-shift mechanism within the augmentation database, which includes temporal bias correction and dual-scorer alignment. Finally, we analyze the TraceBias algorithm and its role in optimizing policies within the ARF-RLHF pipeline.

\subsection{Adaptive Reward-Following (ARF) Scorer}

Studies show that human communication conveys explicit meaning while also signaling satisfaction and willingness to continue~\cite{Chen2016ImplicitPA,book,henry2021willingness,prabhumoye2017linguistic}. Based on this, we introduce two complementary scorers: a static satisfaction scorer that collects quality estimates, and an ARF scorer that updates periodically to track evolving preferences.

\subsubsection{Static Satisfaction Scorer}

Both scorers use the lightweight RoBERTa-mini~\cite{liu2019robertarobustlyoptimizedbert} architecture, which offers low latency and solid semantic understanding.

For self-supervised reward modeling, the static scorer predicts the quality of a (prompt, response) pair from the user’s follow-up reply. Specifically, it takes the follow-up as input and outputs a satisfaction score reflecting the sentiment toward the previous system response.

We project the final hidden states of RoBERTa-mini to three sentiment classes: \textit{negative}, \textit{neutral}, and \textit{positive}, and aggregate token-level logits to obtain a sequence-level satisfaction distribution:

\begin{equation}
    \mathcal{C}_3 = \text{Linear}(\mathcal{H}_{\text{Last}})
\end{equation}
\begin{equation}
\label{NormC}
    \mathcal{C}_{\text{avg}} = \text{Softmax} \left( \frac{1}{L} \sum_{n=1}^{L} \mathcal{C}_3^{(n)} \right)
\end{equation}

Here, $\mathcal{H}_{\text{Last}}$ denotes the final hidden states from RoBERTa-mini, and $\mathcal{C}_3 \in \mathbb{R}^{L \times 3}$ represents token-level satisfaction logits, where $L$ is the input sequence length. The final prediction $\mathcal{C}_{\text{avg}} \in \mathbb{R}^{3}$ summarizes sequence-level satisfaction through mean pooling followed by softmax normalization. The importance of this three-class structure is further discussed in Discussion~\ref{disc:Satisfaction}.

These static predictions are collected as soft labels to train the ARF scorer, which learns to assign reward scores to collected or new (prompt, response) pairs offline. The ARF scorer then serves as the reward function in TraceBias, guiding LLM fine-tuning without manual annotations.

\subsubsection{ARF Scorer}

The ARF scorer is fine-tuned during interactions to adapt to changing preferences. At each training step, it outputs a predicted satisfaction distribution $\hat{\mathcal{C}}$ for LLM feedback. The prediction is supervised using soft labels derived from user feedback and historical interactions.

To speed up convergence while keeping baseline estimates stable, the ARF scorer builds on the static satisfaction scorer. The averaged satisfaction vector $\mathcal{C}_{\text{avg}}$, distilled from user follow-up messages, provides soft guidance for supervision. We employ the standard cross-entropy loss:

\begin{equation}
    \mathcal{L}_{\text{supervised}} = \text{CE}(\hat{\mathcal{C}}, \mathcal{C}_{\text{avg}})
\end{equation}

To mitigate overfitting and catastrophic forgetting when real-time data is scarce, we use an Experience Replay (ER) mechanism. A sampling ratio $\text{ER}_{\text{ratio}}$ switches training probabilistically between historical data and current feedback:

\[
\mathcal{L}_{\text{total}} =
\begin{cases}
    \mathcal{L}_{\text{ER}} = \text{CE}(\hat{\mathcal{C}}, \mathcal{C}_{\text{static}}), & \text{if } p < \text{ER}_{\text{ratio}} \\
    \mathcal{L}_{\text{supervised}} = \text{CE}(\hat{\mathcal{C}}, \mathcal{C}_{\text{avg}}), & \text{otherwise}
\end{cases}
\]

Here, $\mathcal{C}_{\text{static}}$ denotes labels from the static satisfaction dataset (e.g. DailyDialog~\cite{li-etal-2017-dailydialog}, GoEmotions~\cite{demszky2020goemotionsdatasetfinegrainedemotions}), and $p$ is a random variable, sampled from a uniform distribution $p \sim \text{Uniform}(0,1)$. This alternating strategy lets the ARF scorer benefit from both stable historical signals and dynamic user feedback, improving generalization and robustness.

\subsection{Augmentation Database}
For better use limited user feedback, we build an Augmentation Database that expands training data with synonym substitution and random trace truncation. We use a score-shift mechanism, blending static and adaptive distributions, ensuring augmented samples align with changing preferences.

\subsubsection{Preference-Biased Fusion Mechanism}

To ensure robust and adaptive scoring, we unify the preference-biased update into a single mechanism. 
Given two candidate distributions $\mathcal{C}_A$ and $\mathcal{C}_B$, we compute a dynamic weighting coefficient:
\begin{equation}
    \mathcal{S}_{\text{cos}} = \sigma \left( \big(\text{CosSim}(\mathcal{C}_A, \mathcal{C}_B) - 0.5 \big) \cdot S_{\text{sig}} \right),
\end{equation}
where $\sigma$ denotes the sigmoid, $\text{CosSim}$ the cosine similarity, and $S_{\text{sig}}$ a sensitivity scaling factor (Appendix~\ref{app:Scale} for selection). The fused score is then:
\begin{equation}
    \mathcal{C}_{\text{fused}} = \mathcal{C}_A \cdot \mathcal{S}_{\text{cos}} + \mathcal{C}_B \cdot (1 - \mathcal{S}_{\text{cos}})
\end{equation}

This covers two cases:
\begin{compactitem}
    \item \textbf{Augmented Data Scoring:}\\
    $\mathcal{C}_A = \mathcal{C}_{\text{ARF\_avg}}, \quad 
      \mathcal{C}_B = \mathcal{C}_{\text{basic\_avg}}$\\
    \textit{$\mathcal{C}_{\text{ARF\_avg}}$: mean score from the adaptive ARF scorer.}\\
    \textit{$\mathcal{C}_{\text{basic\_avg}}$: mean score from the static scorer.}

    \item \textbf{Historical Re-Scoring:}\\
    $\mathcal{C}_A = \mathcal{C}_{\text{new\_avg}}, \quad 
      \mathcal{C}_B = \mathcal{C}_{\text{old\_avg}}$\\
    \textit{$\mathcal{C}_{\text{new\_avg}}$: mean score from the updated ARF scorer.}\\
    \textit{$\mathcal{C}_{\text{old\_avg}}$: mean score stored from past evaluations.}
\end{compactitem}

This balances stability from static or historical references with adaptability to evolving preferences, preventing overfitting while maintaining continuity.

\subsection{TraceBias Algorithm}

Given enough preference-rich interaction data, we optimize policies with a score-based actor–critic algorithm called \textbf{TraceBias}. While theoretically aligned with PPO and DPO (see Appendix~\ref{app:TraceBias-AC}), TraceBias uses a token-wise strategy and new normalization methods for stable, fine-grained optimization, without binary comparisons (see pseudocode in Appendix~\ref{app:pseudocode}).

The algorithm has two key components: (1) a \textbf{Double Average Method (DAM)} that normalizes satisfaction scores and token-level policy ratios, stabilizing training across variable-length sequences; and (2) an advantage function derived from trajectory-level score differences between fine-tuned and reference models, serving as the main optimization signal.

\subsubsection{Double Average Method (DAM)}
\label{DAM}

We address gradient inconsistencies that arise from variable sequence lengths. Rather than using explicit gradient clipping—which can remove informative signals—we normalize both satisfaction scores and token-level policy ratios.

Let \(T_i\) be the sequence length (number of tokens in an output), 
\(N\) the number of turns in a trajectory, 
and \(i \in \{1, \dots, N\}\) the turn index. 
Denote \(x^{(i)}\) as the input at turn \(i\), and \(y_t^{(i)}\) as its \(t\)-th generated token. 
\(P_{\theta}\) and \(P_{\text{old}}\) are the current and reference policy probabilities. 
We define the average log-probabilities under each policy and their ratio as:

\begin{equation}
\pi_{\theta}^{(i)} = \frac{1}{T_i} \sum_{t=1}^{T_i} \log P_{\theta}(y_t^{(i)} \mid x^{(i)})
\end{equation}
\begin{equation}
\pi_{\text{old}}^{(i)} = \frac{1}{T_i} \sum_{t=1}^{T_i} \log P_{\text{old}}(y_t^{(i)} \mid x^{(i)})
\end{equation}
\begin{equation}
\label{eq:ratio}
\pi_{\text{ratio}}^{(i)} = \exp(\pi_{\theta}^{(i)} - \pi_{\text{old}}^{(i)})
\end{equation}

By combining satisfaction score normalization $\mathcal{C}_{\text{avg}}$  (Eq.~\ref{NormC}) with normalized token-level ratios, DAM reduces gradient imbalance—where long sequences otherwise dominate updates and short ones diminish. This produces more stable gradients than direct clipping, as illustrated in Fig.~\ref{fig:GradCompare}.

\subsubsection{Trace Scores with Discounted Step-wise Evaluation}

With token-level optimization stabilized, we compute score differences between generated and reference trajectories—\(\mathcal{S}_{\text{t-real}}\) and \(\mathcal{S}_{\text{t-ref}}\)—to estimate the advantage function.

At each evaluation step \(j \in \{1, \dots, i\}\) within the first \(i\) turns, 
we define relative preference scores \(\mathcal{S}^{(j)}_{\text{real}}\) and \(\mathcal{S}^{(j)}_{\text{ref}}\), 
corresponding to the currently generated response and a previously generated response retrieved from the augmentation database, respectively. 
Both are computed via:
\begin{equation}
\label{eq:neu}
    \mathcal{S}^{(j)} = \mathcal{C}_{\text{avg[2]}}^{(j)} - \mathcal{C}_{\text{avg[0]}}^{(j)}
\end{equation}
where \(\mathcal{C}_{\text{avg[2]}}^{(j)}\) and \(\mathcal{C}_{\text{avg[0]}}^{(j)}\) represent positive and negative sentiment, respectively.

To aggregate over a multi-turn trajectory, we apply a discount factor \(\gamma\):
\begin{equation}
    \mathcal{S}^{(i)}_{\text{t-real}} = \sum_{j=1}^{i} \gamma^{j-1} \cdot \mathcal{S}^{(j)}_{\text{real}}, \quad
    \mathcal{S}^{(i)}_{\text{t-ref}} = \sum_{j=1}^{i} \gamma^{j-1} \cdot \mathcal{S}^{(j)}_{\text{ref}}
\end{equation}

The sentiment gap $\mathcal{S}_{\text{t-ref}}^{(i)} - \mathcal{S}_{\text{t-real}}^{(i)}$ then acts as the advantage signal, allowing TraceBias to capture evolving sentiment and weight dialogue turns unequally.

\subsubsection{Final Representation of TraceBias}

TraceBias combines trajectory-level signals with DAM-normalized policy dynamics to define its training objective. Integrating the trajectory-based advantages with normalized ratios yields:

\begin{equation}
\label{eq:final}
    \mathcal{L}_{\text{final}} = -\sum_{i=1}^{N} \left( \mathcal{S}_{\text{t-ref}}^{(i)} - \mathcal{S}_{\text{t-real}}^{(i)} \right) \cdot \pi_{\text{ratio}}^{(i)}
\end{equation}

This objective enables stable, fine-grained updates without relying on binary preference labels, forming a robust optimization basis for the broader ARF framework.

\section{Experiments}

\subsection{Experimental Setup}
We systematically evaluate our proposed \textbf{ARF-RLHF} framework against widely used baselines, including \textbf{DPO}~\cite{rafailov2024directpreferenceoptimizationlanguage}, \textbf{PPO}~\cite{schulman2017proximalpolicyoptimizationalgorithms}, \textbf{RLAIF}~\cite{lee2024rlaif}, and \textbf{KTO}~\cite{ethayarajh2024ktomodelalignmentprospect}.  
All models are fine-tuned using \textbf{LoRA}~\cite{hu2022lora}, which provides parameter-efficient adaptation without compromising performance.  

Our evaluation spans four representative lightweight LLMs—\textbf{Gemma2-2B}~\cite{gemmateam2024gemma2improvingopen}, \textbf{Qwen2-1.5B}~\cite{yang2024qwen2technicalreport}, \textbf{Qwen2.5-1.5B}~\cite{qwen2025qwen25technicalreport}, and \textbf{LLaMA3.2-3B}~\cite{grattafiori2024llama3herdmodels}—ensuring that findings are not specific to a single architecture or training recipe. We further cover five diverse datasets from Big-Bench~\cite{srivastava2023beyond}, targeting different capabilities: \textbf{Alpaca} (instruction following), \textbf{GSM8K} (mathematical reasoning), \textbf{StrategyQA} (commonsense reasoning), \textbf{TopicalChat} (dialogue), and \textbf{CNN/DailyMail} (summarization). This combination allows us to test both reasoning-heavy and open-domain conversational scenarios. In addition, we include evaluation on the \textbf{HH-RLHF} human-annotated dataset~\cite{bai2022traininghelpfulharmlessassistant}, which serves as a high-quality benchmark for alignment under real human feedback and enables direct comparison with prior studies such as KTO.

Additionally, we introduce a large-scale preference dataset, \textbf{Emotion3}, which integrates 78,630 samples from DailyDialog, GoEmotions, ISEAR~\cite{isear}, and Sentiment140~\cite{go2009twitter}.  
Compared with typical RLHF datasets that are small and domain-limited, Emotion3 is deliberately broad and noisy, reflecting the variability of real-world user feedback.

All details of dataset construction, validation, and experimental hyperparameter settings are provided in Appendix~\ref{app:Satisfaction Dataset} and Appendix~\ref{app:hyper}.

This experimental design enables us to probe the following three questions:  
\begin{compactenum}
    \item[\textbf{Q1}] Can ARF reliably collect and adapt to dynamic user preferences, especially under noisy or shifting conditions?  
    \item[\textbf{Q2}] How does TraceBias compare with PPO, DPO, RLAIF and KTO when evaluated under consistent and fair reward supervision?  
    \item[\textbf{Q3}] Which specific mechanisms in ARF-RLHF (e.g., ER, DAM, rescoring) are essential for stability and generalization?  
\end{compactenum}

\subsection{Q1: Can ARF Reliably Track Preferences?}

A key challenge in RLHF is whether feedback signals can be both \textbf{accurate} and \textbf{adaptable} over time.  
Static scorers provide stable supervision but may fail once user preferences shift, whereas dynamic scoring risks instability.  
We therefore first test ARF’s reliability in both static and dynamic scenarios.  

\paragraph{Static supervision quality.}  
Table~\ref{tab:static} reports the accuracy of our static satisfaction scorer across five sentiment benchmarks.  
The consistently high accuracy (all $>$70\%) demonstrates that ARF can rely on this module as a trustworthy backbone.  
This step is crucial: without verifying the base quality of preference signals, subsequent alignment results would be difficult to interpret.

\begin{table}[tb]
    \centering
    \small
    \resizebox{\columnwidth}{!}{
    \begin{tabular}{lccccc}
    \toprule
    \textbf{Dataset} & DailyDialog & GoEmotions & ISEAR & Sentiment140 & Emotion3 \\
    \midrule
    Accuracy (\%) & 70.05 & 73.65 & 76.00 & 74.10 & 71.60 \\
    \bottomrule
    \end{tabular}}
    \caption{Test accuracy of the static satisfaction scorer. Reliable static signals are a prerequisite for effective RLHF.}
    \label{tab:static}
\end{table}

\begin{figure}[tb]
    \centering
    \includegraphics[width=\columnwidth]{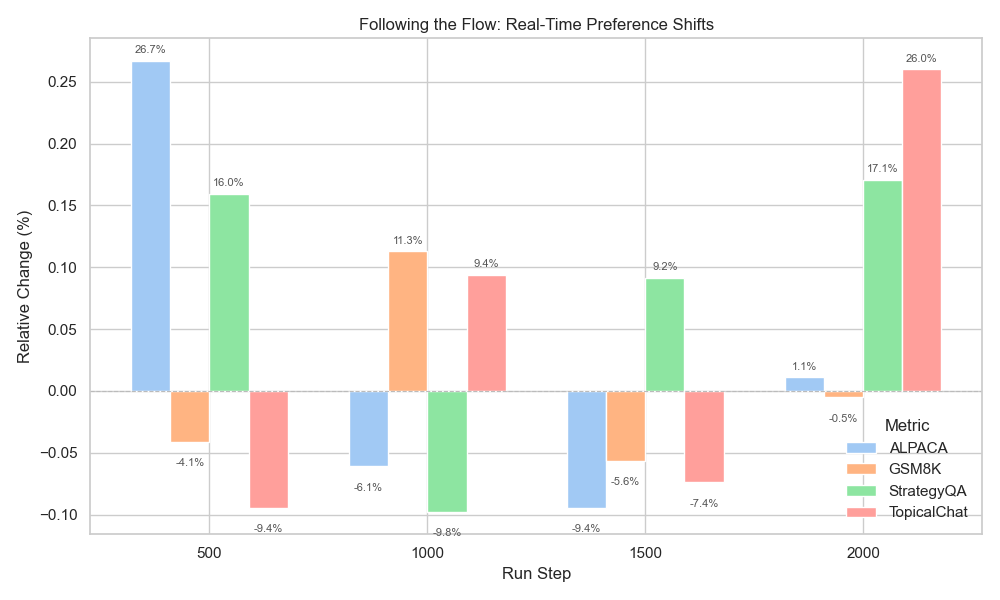}
    \caption{Tracking preference shifts using ARF. Performance drops reflect deliberate adaptation to new negative signals, validating robustness under non-stationary feedback.}
    \label{fig:ARFTracking}
    \vspace{-1em}
\end{figure}

\paragraph{Adaptive preference tracking.}  
We then examine whether ARF can dynamically adapt to shifting preferences by sequentially injecting bias‑specific data every 500 steps (ALPACA $\rightarrow$ GSM8K $\rightarrow$ StrategyQA $\rightarrow$ TopicalChat), leveraging the static scorer’s accuracy as a reference. When more than two biases overlap, negative supervision is applied to the earliest bias. As shown in Figure~\ref{fig:ARFTracking}, ARF successfully tracks these shifts, with clear downward adjustments once negative preferences are introduced. Initially, synchronized gain/loss patterns emerge between ALPACA and StrategyQA, and between GSM8K and TopicalChat, likely due to semantic similarity; nevertheless, ARF rapidly disentangles these correlations once negative supervision is applied (e.g., ALPACA at step 1500), avoiding stale alignment and demonstrating robustness to preference contradictions—a scenario often overlooked in prior RLHF research.

\subsection{Q2: How Does TraceBias Compare Against RLHF Baselines?}
\label{Q2}

Most prior work evaluates PPO and DPO using multiple LLM judges, but these methods are sensitive to prompt phrasing and evaluation bias.  
To ensure fairness, we employ a \textbf{unified reward model} both for preference filtering and performance evaluation (See more in discussion~\ref{dic:eval}).  
This allows us to ask: under identical supervision, which method learns most effectively?  

\begin{figure}[tb]
    \centering
    \includegraphics[width=\columnwidth]{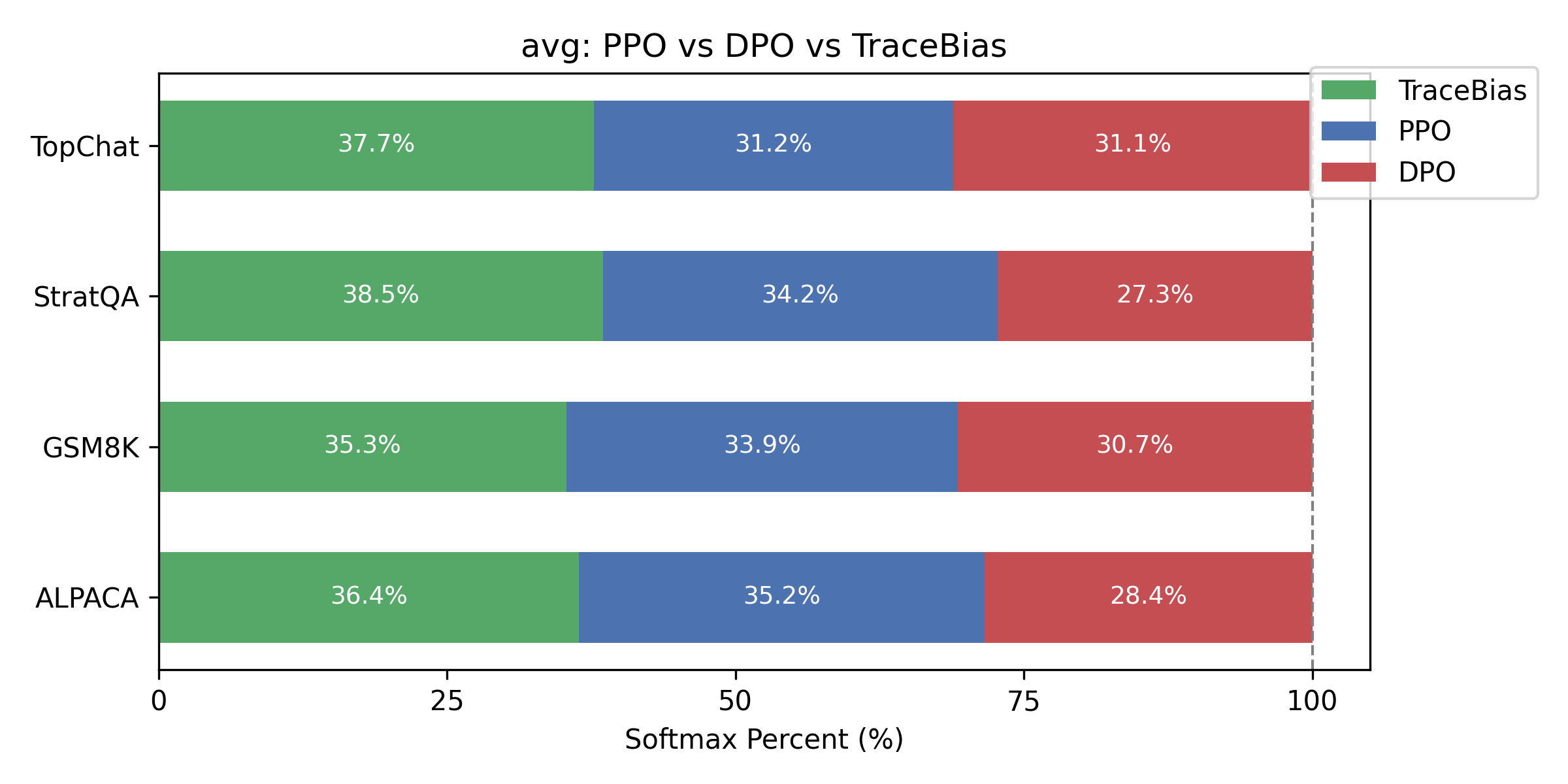}
    \caption{Average performance comparison under different baselines' fine-tuning. TraceBias consistently outperforms PPO and DPO across tasks. Single models' performance in appendix~\ref{app:RLHFBaselines}.}
    \label{fig:avgRLHF}
    \vspace{-1em}
\end{figure}

\paragraph{Performance across tasks.}  
Figure~\ref{fig:avgRLHF} shows that TraceBias consistently outperforms both PPO and DPO across all tasks and model sizes, with average improvements of +3.3\% over PPO and +7.6\% over DPO.  
These gains are not marginal: in low-resource RLHF settings, even small percentage improvements can determine whether the aligned model is useful in practice.  
We attribute the improvements primarily to \textbf{Double Average Method (DAM)}, which stabilizes training updates and prevents reward hacking.

\paragraph{Robustness under LLM-based preferences.}  
\label{exp:RLAIF}
Human preference data are expensive and often limited, making synthetic labels an appealing alternative. We construct a 1K StrategyQA preference dataset using DeepSeek‑v3~\cite{deepseek}. As shown in Table~\ref{tab:RLAIF}, TraceBias outperforms both RLAIF‑PPO and RLAIF‑DPO under this synthetic supervision. Interestingly, DPO performs slightly better than PPO here, likely because the preference signals are clearer—suggesting it benefits from more finely curated datasets. Together with the results in the \textit{performance across tasks} section, these findings show that TraceBias delivers superior robustness, consistently maintaining strong performance across diverse bias conditions and supervision regimes.

\begin{table}[tb]
    \centering
    \small
    \resizebox{\columnwidth}{!}{
    \begin{tabular}{lccc}
    \toprule
    Evaluation method & RLAIF-PPO & RLAIF-DPO & TraceBias \\
    \midrule
    Preference Score (\%) & 30.3 & 32.8 & 36.9 \\
    DeepSeek-V3 (win rate vs.\ Base) & 43\% & 49\% & 52\% \\
    \bottomrule
    \end{tabular}}
    \caption{Comparison of RLAIF variants and TraceBias on StrategyQA with DeepSeek-v3 preferences. TraceBias achieves the highest win rate, confirming robustness under noisy labels.}
    \label{tab:RLAIF}
\end{table}

\paragraph{Evaluation on human-annotated data.}  
\label{exp:HH-RLHF}
To further verify robustness in realistic scenarios, we employ a hybrid validation strategy that combines human annotations with LLM-based judgments from GPT-5~\cite{openai2025gpt5systemcard}, confirming that the unified scorer remains a reliable basis for evaluation. 
Based on this setup, we additionally evaluate DPO, KTO, and TraceBias on a real human-labeled dataset.  
As shown in Table~\ref{tab:HH-RLHF}, TraceBias again achieves competitive performance, slightly outperforming KTO and closely matching DPO.  
Notably, DPO shows stronger performance under high-quality human annotations, which is consistent with prior findings in the KTO paper (Section 4.5: \textit{KTO vs.\ DPO – when to use which?}).  
This observation aligns with our earlier discovery: while KTO is more robust under noisy or synthetic preference labels, DPO tends to excel when reliable human feedback is available, and TraceBias balances both regimes effectively.  
Importantly, even when all three methods are evaluated on the same human-annotated dataset, their relative advantages remain consistent.

\begin{table}[tb]
    \centering
    \small
    \resizebox{\columnwidth}{!}{
    \begin{tabular}{lccc}
    \toprule
    Evaluation method & DPO & KTO & TraceBias \\
    \midrule
    Preference Score (\%) & 34.2 & 31.7 & 34.1 \\
    GPT-5 (win rate vs.\ Base) & 57\% & 51\% & 54\% \\ 
    \bottomrule
    \end{tabular}}
    \caption{Comparison of DPO, KTO, and TraceBias on a real human-annotated dataset HH-RLHF. }
    \label{tab:HH-RLHF}
\end{table}

\subsection{Q3: Which Design Components Are Critical?}

While ARF-RLHF introduces several innovations, it is important to understand which components are indispensable.  
We therefore ablate \textbf{Experience Replay (ER)}, \textbf{Double Average Method (DAM)}, and \textbf{rescoring}.  

\paragraph{Experience Replay (ER).}  
Table~\ref{tab:ER} shows that removing ER (ratio = 0) yields higher short-term accuracy on recent tasks but significantly degrades performance on the Emotion3 dataset, indicating poor generalization.  
Conversely, a moderate ER ratio (0.5) balances adaptation and long-term robustness.  
This suggests that ER acts as a memory mechanism, mitigating catastrophic forgetting when preferences evolve—a property often overlooked in existing RLHF pipelines.  

\begin{table}[tb]
    \centering
    \small
    \resizebox{\columnwidth}{!}{
    \begin{tabular}{lcc}
    \toprule
    ER Ratio & GSM8K (accuracy) & Emotion3 (accuracy) \\
    \midrule
    Basic & 53.52 & 73.84 \\
    0     & 60.59 & 59.32 \\
    0.5   & 56.40 & 70.88 \\
    \bottomrule
    \end{tabular}}
    \caption{Ablation of ER ratio. A moderate setting improves generalization and yields higher accuracy, whereas no replay causes catastrophic forgetting with a sharp drop in accuracy.}
    \label{tab:ER}
\end{table}

\paragraph{Gradient stability.}  
Figure~\ref{fig:GradCompare} compares gradient dynamics on Qwen2.5-1.5B fine-tuned over GSM8K.  
While PPO suffers from high variance even with clipping, TraceBias with DAM yields both lower magnitude and variance.  
Stable gradients are crucial in RLHF, where unstable updates can quickly derail alignment.  
This validates DAM as a key factor behind TraceBias’s performance.  

\paragraph{Rescoring necessity.} As shown earlier in Table~\ref{tab:ReScore}, the ablation study compares the use and omission of ReScoring within experiment in \textit{Adaptive Preference Tracking} via ARF. When ALPACA and GSM8K undergo preference reversal, removing rescoring results in inflated scores and persistent misalignment. This demonstrates that rescoring is not merely an auxiliary option but an essential mechanism to ensure preference signals remain valid over time. 

\subsection{Summary of Findings}
In summary, our experiments show that ARF reliably captures preference signals and adapts to evolving supervision even under noise or conflicts, TraceBias matches the performance of PPO and DPO across both human-annotated and synthetic preferences, and core components such as ER, DAM, and rescoring are indispensable for stable training and long-term alignment. Detailed case studies can be found in Appendix~\ref{app:CaseStudy}.

\section{Discussion}

\subsection{How can we prove the accuracy of our experiments?}
\label{dic:eval}

As noted in Section~\ref{Q2}, AI‑agent–based evaluation (e.g., using an LLM judge) exhibits high variance from prompt wording, task quirks, model architecture, and random seeds, leading to inconsistent results. We therefore focus on comparing the convergence of different RLHF methods under identical preference supply. To mitigate variance and ensure real‑world applicability, we combine AI‑agent metrics (Section~\ref{exp:RLAIF}) with a unified scorer: reward‑oriented methods (e.g., TraceBias, PPO) train against a fixed reward model, while comparison‑based methods (e.g., DPO, KTO) use the same scorer for alignment. All runs share the same held‑out test set (unseen in training) and immutable scorer, preventing coupling or leakage. In RLAIF and HH‑RLHF experiments, we further employ two distinct AI‑agent evaluations (DeepSeek-V3 and GPT‑5); though less reproducible, they provide complementary evidence that our scorer remains valid in realistic settings, enabling stable, unbiased comparison of convergence to target preferences.

\begin{table}[tb]
    \centering
    \small
    \resizebox{\columnwidth}{!}{
    \begin{tabular}{lcc}
    \toprule
    \textbf{Condition} & \textbf{ALPACA ($\Delta$\%)} & \textbf{GSM8K ($\Delta$\%)} \\
    \midrule
    With ReScore & -9.4 & -0.5 \\
    Without ReScore & 7.8 & 3.7 \\
    \bottomrule
    \end{tabular}}
    \caption{Relative change ($\Delta$\%) on ALPACA and GSM8K after preference reversal. Negative values indicate successful adaptation.}
    \label{tab:ReScore}
\end{table}

\subsection{On the Reliability of ARF Satisfaction Supervision}
\label{disc:Satisfaction}

Satisfaction annotations are inherently subjective-labeling tasks involving large numbers of human annotators often reflect diverse preferences, even SOTA LLMs show bias when re-labeling Emotion3, with neutral predictions ranging from 24.0\% to 37.3\% (variance 29.51, Appendix~\ref{app:Subjective}). Neutrality proves especially ambiguous, leading to unstable supervision.  Although our static scorer reaches only ~70\% accuracy, it mitigates such uncertainty by excluding neutral scores during TraceBias updates (Eq.~\ref{eq:neu}), using them instead to dampen noisy gradients e.g., Good: 0.02, Neutral: 0.90, Bad: 0.08 yields \(S = -0.06\) resulting in minimal updates. This design—by ensuring softer updates for items dominated by neutral evaluations—avoids rigid binary comparisons under ambiguity, thereby improving robustness, reducing annotation variance, and enabling reliable tracking of individual preferences beyond crowd-level bias.

\section{Conclusion}
We introduce \textbf{ARF-RLHF}, a reinforcement learning framework that aligns language models to individual user preferences. By modeling continuous satisfaction and adapting to evolving feedback, it enables personalized, context-aware behavior beyond population-level signals. Experiments show robust preference optimization under limited supervision, offering a scalable path to adaptive models that reflect individual intent while maintaining rigor and reliability.

\section{Limitations}
While our method offers significant theoretical and empirical advantages, we acknowledge the following limitations:

\begin{compactitem}
\item \textbf{Lack of Real Human evaluation}: The scorer‑based evaluation protocol is designed for fair comparison of RLHF methods, focusing on minimizing noise factors. It does not yet capture generalization performance in real-world scenarios. Future work will include human blind testing or cross-validation with alternative scorers to further validate the results.

\item \textbf{Model Scale Constraint}: Due to resource constraints, we conduct experiments on 1.5B–3B models. While our method is architecture-agnostic and theoretically scalable, its effectiveness on larger LLMs (e.g., 7B, 13B, 65B) remains to be explored in future work.
\end{compactitem}
\bibliography{custom}

\appendix

\section{Hyperparameters}
\label{app:hyper}
All experiments' Hyperparameters shows below:
\begin{enumerate}
    \item Static Scorer Evaluation: table~\ref{hyp:Static}
    \item Adaptive Preference Tracking: table~\ref{hyp:ARF tracking}
    \item RLHF Comparison: table~\ref{hyp:RLHF}
    \item  RLAIF vs. ARF: table~\ref{hyp:RLAIF}
    \item Effect of Experience Replay (ER) in ARF~\ref{hyp:ER}
    \item Gradient Stability Analysis for DAM~\ref{hyp:DAM}
\end{enumerate}
All experiments were conducted on 2 * NVIDIA GTX 2080 Ti GPUs with CUDA unified memory enabled, using multi-GPU parallelism. Training consumed approximately 390 GPU-hours per device.

\section{The Essential Homology of DPO, PPO, and TraceBias}
\label{app:Homology}
The Actor-Critic (AC) algorithm can be formulated as:
\begin{equation}\label{AC}
\mathcal{L}^{\text{AC}}(\theta) = - \sum_{t=1}^{N} \log \pi_\theta(a_t \mid s_t) \cdot A_t
\end{equation}
In this section, we demonstrate the theoretical connections among PPO, DPO, and our proposed TraceBias. We argue that these methods share a common optimization structure, thereby validating the theoretical soundness of TraceBias.

\subsection{PPO as a Clip-Constrained Actor-Critic Variant}

PPO~\cite{schulman2017proximalpolicyoptimizationalgorithms} can be defined as:
\begin{equation}
\mathcal{L}^{\text{PPO}}(\theta) = \mathbb{E}_t \left[ \min \left( r_t(\theta) A_t,\ \text{clip}(r_t(\theta), \epsilon) A_t \right) \right]
\end{equation}

where $r_t(\theta)$ is the importance sampling ratio between the current and previous policies:
\begin{equation}
r_t(\theta) = \frac{\pi_\theta(a_t \mid s_t)}{\pi_{\theta_{\text{old}}}(a_t \mid s_t)}
\end{equation}
and the clipping function is defined as:
\begin{equation}
\text{clip}(r_t(\theta),\epsilon) =
\begin{cases}
1 - \epsilon, & \text{if } r_t(\theta) < 1 - \epsilon \\
r_t(\theta), & \text{if } 1 - \epsilon \leq r_t(\theta) \leq 1 + \epsilon \\
1 + \epsilon, & \text{if } r_t(\theta) > 1 + \epsilon
\end{cases}
\end{equation}

By expanding the objective, we obtain:
\begin{align}
\mathcal{L}^{\text{PPO}}(\theta) =
\mathbb{E}_t \Bigg[
\min \Bigg(
& \frac{\pi_\theta(a_t \mid s_t)}{\pi_{\theta_{\text{old}}}(a_t \mid s_t)}A_t ,\nonumber 
 \text{clip}\\
& \left(
\frac{\pi_\theta(a_t \mid s_t)}{\pi_{\theta_{\text{old}}}(a_t \mid s_t)},\epsilon 
\right) A_t
\Bigg)
\Bigg]
\end{align}

If we ignore the clipping operation—which serves as a regularization mechanism to limit the policy update magnitude—the PPO objective reduces to:
\begin{equation}
\mathcal{L}^{\text{PPO}}(\theta) \propto \sum_{t=1}^{N} r_t(\theta) A_t
\end{equation}
Here, $r_t(\theta)$ reflects the policy ratio $ \frac{\pi_\theta(a_t \mid s_t)}{\pi_{\theta_{\text{old}}}(a_t \mid s_t)}$, which encourages increasing the likelihood of actions with high advantage $A_t$. This shows that PPO essentially shares the same optimization direction as AC, differing only in the incorporation of a trust-region-inspired constraint to stabilize training.

\subsection{DPO as a KL-Constrained Actor-Critic Variant}

Direct Preference Optimization (DPO)~\cite{rafailov2024directpreferenceoptimizationlanguage} leverages the Bradley-Terry model to represent pairwise preferences as follows:
\begin{equation}
\mathbb{P}(y^+ \succ y^- \mid x) = \frac{ \exp(r(y^+)) }{ \exp(r(y^+)) + \exp(r(y^-)) }
\end{equation}

Its associated loss is:

\begin{align}
\mathcal{L}_R(\phi, D)
&= - \mathbb{E}_{(x,\,y_w,\,y_l)} \notag \\
& \left[ 
\log \frac{ \exp(r(y_w)) }{ \exp(r(y_w)) + \exp(r(y_l)) }
\right]
\end{align}

The DPO objective derived from this model is:
\begin{align}
\mathcal{L}_{\mathrm{DPO}}(\pi_\theta;\,\pi_{\mathrm{ref}})
&= - \mathbb{E}_{(x,y_w,y_l)} \left[ \right. \notag\log  \\
& \sigma\left.
 \left(
\beta \log \frac{
\pi_\theta(y_w\mid x)\, \pi_{\mathrm{ref}}(y_l\mid x)
}{
\pi_\theta(y_l\mid x)\, \pi_{\mathrm{ref}}(y_w\mid x)
}
\right)
\right]
\end{align}

Introducing a normalization constant $Z(x)$, the implicit optimal policy $\pi^*(y \mid x)$ can be defined as:
\begin{equation}
\pi^*(y \mid x) = \frac{ \pi_{\text{ref}}(y \mid x) \cdot \exp(\frac{1}{\beta} r(y)) }{ Z(x) }
\end{equation}
where the partition function $Z(x)$ is:
\begin{equation}
Z(x) = \sum_{y'} \pi_{\text{ref}}(y' \mid x) \cdot \exp\left(\frac{1}{\beta} r(y')\right)
\end{equation}

Taking the logarithm of both sides yields:
\begin{equation}
\log \pi^*(y \mid x) = \log \pi_{\text{ref}}(y \mid x) + \frac{1}{\beta} r(x,y) - \log Z(x)
\end{equation}

We can then derive:
\begin{equation}
r(x, y) = \beta \log \frac{\pi^*(y \mid x)}{\pi_{\mathrm{ref}}(y \mid x)} + \beta \log Z(x)
\end{equation}

By applying $r(x, y)$ in the pairwise preference model $\mathbb{P}(y^+ \succ y^- \mid x)$ and utilizing the Plackett-Luce model (see Appendix A.3 of~\cite{rafailov2024directpreferenceoptimizationlanguage} for more details), the DPO objective can be equivalently rewritten as:
\begin{equation}
\begin{split}
\max_{\pi_\theta}\;\Bigl\{\, 
&\mathbb{E}_{x\sim D,\,y\sim\pi_\theta(\cdot\mid x)}\bigl[r_\phi(x,y)\bigr] \\ 
&\quad -\,\beta\,\mathrm{KL}\!\bigl(\pi_\theta(\cdot\mid x)\,\|\,\pi_{\mathrm{ref}}(\cdot\mid x)\bigr)
\Bigr\}
\end{split}
\end{equation}

Here, the optimization objective is to maximize the expected reward regularized by a KL divergence term. Assuming $A(x, y) = r(x, y)$, and temporarily ignoring the KL regularization, this reduces to an actor-critic style objective:
\begin{equation}
\mathcal{L}^{\text{DPO}}(\theta) \propto \mathbb{E}_{y \sim \pi_\theta(\cdot \mid x)} \left[ A(x, y) \right]
\end{equation}

This implies that DPO and actor-critic share effectively the same optimization objective when the reward signal is defined as the advantage. In practice, this expectation can be approximated by:
\begin{equation}
\mathcal{L}^{\text{DPO}}(\theta) \approx \sum_{i=1}^{N} \log \pi_\theta(y_i \mid x_i) \cdot A(x_i, y_i)
\end{equation}

Thus, DPO can be interpreted as a KL-regularized actor-critic method, where the reward signal is derived from preference feedback rather than scalar returns.

\subsection{TraceBias as a DAM-Constrained Actor-Critic Variant}
\label{app:TraceBias-AC}

We have previously outlined the Actor-Critic (AC) interpretation of TraceBias in our methodology. Here, we provide a formal derivation from the expanded formulation to its AC-style representation.

\begin{equation}
\begin{split}
    \mathcal{L}_{\text{final}} = -\sum_{i=1}^{T} 
    &\left( 
        \mathcal{S}_{\text{t-ref}}
        - 
        \mathcal{S}_{\text{t-real}}
    \right) \cdot \exp(\pi_{\theta}^{(i)} - \pi_{old}^{(i)})
\end{split}
\end{equation}

Following our methodology, the token-level reward (or \emph{score}) is defined via discounted($\gamma$) step-wise preferences as follows:
\begin{equation}
    \mathcal{S}_{\text{t-real}}^{(i)} = \sum_{j=1}^{i} \gamma^{j-1} \cdot \mathcal{S}^{(j)}_{\text{real}}, \quad
    \mathcal{S}_{\text{t-ref}}^{(i)} = \sum_{j=1}^{i} \gamma^{j-1} \cdot \mathcal{S}^{(j)}_{\text{ref}}
\end{equation}

Accordingly, we define the advantage function by measuring the difference between the real and reference trajectories:
\begin{equation}
    \mathcal{A}_i = \mathcal{S}_{\text{t-ref}}^{(i)} - \mathcal{S}_{\text{t-real}}^{(i)}
\end{equation}

Substituting this into the objective, TraceBias can be rewritten in an actor-critic form:
\begin{equation}
\begin{split}
\mathcal{L}_{\mathrm{TraceBias}}
&= -\sum_{i=1}^{N} \mathcal{A}_i \cdot
   \exp(\pi_{\theta}^{(i)} - \pi_{old}^{(i)})\\[-0.5ex]
\end{split}
\end{equation}

To improve optimization stability, we introduce the DAM smooth surrogate strategy, which pools token-level scores and normalizes the policy ratio across the trajectory. Analogous to the clipping term in PPO and the KL regularization in DPO, DAM serves as a regularization mechanism:
We define the token-level ratio as:
\begin{equation}
\pi_{\theta}^{(i)} = \frac{1}{T_i} \sum_{t=1}^{T_i} \log P_{\theta}(y_t^{(i)} \mid x^{(i)})
\end{equation}

\begin{equation}
\pi_{\text{old}}^{(i)} = \frac{1}{T_i} \sum_{t=1}^{T_i} \log P_{\text{old}}(y_t^{(i)} \mid x^{(i)})
\end{equation}
Then, the normalized trajectory-level policy ratio is computed as:

\begin{equation}
\pi_{\text{ratio}}^{(i)} = \exp(\pi_{\theta}^{(i)} - \pi_{old}^{(i)})
\end{equation}

Thus, the final form of TraceBias is:
\begin{equation}
    \mathcal{L}_{\text{TraceBias}} = -\sum_{i=1}^{T} \mathcal{A}_i
    \cdot \pi_{\text{ratio}}^{(i)}
\end{equation}

This derivation shows that TraceBias can be directly interpreted as an Actor–Critic method without introducing additional approximations, highlighting its theoretically grounded and streamlined formulation.

\subsection{Summary}

Although PPO, DPO, and TraceBias differ significantly in their final objectives, all can be reformulated as variants of the Actor-Critic (AC) framework. By optimizing the expected reward weighted by advantage, each method introduces distinct regularization strategies—such as PPO's clipping, DPO's KL constraint, or TraceBias's DAM normalization—to improve training stability. This unified perspective highlights that TraceBias is not only theoretically grounded but also competitive with existing policy optimization techniques.

\section{The pseudocode of TraceBias}
\label{app:pseudocode}
The TraceBias pseudocode shows in alg~\ref{alg:Trace}.

\section{The Selection of Sigmoid Scale}
\label{app:Scale}
Scale plays a crucial role in determining the sensitivity of the norm function, as it directly affects most parameters. We recommend selecting scales within the range of [4,8], as excessively large values can push most parameters towards the boundaries of the Sigmoid function interval, limiting their effective range.
Table~\ref{tab:Scale} illustrates when $(\text{CosSim}(\mathcal{C}_{\text{ARF\_avg}}, 
    \mathcal{C}_{\text{basic\_avg}}) - 0.5)$ equals values
0.2 is converted together with the corresponding function value range by the Sigmoid function under different proportional settings. It is worth noting that when the scale is set to 1, the relatively large 0.2 level in sigmoid results in a mapping value of only 0.5498. However, when the ratio increases to 16, the same input is mapped to nearly 1, indicating that the range is overly compressed.
Based on these observations, we strongly recommend selecting a scale within the range of [4,8] for the numerical deviation annotation algorithm, as it ensures a balanced transformation without pushing values to extremes.

\begin{table}[tb]
    \centering
    \scriptsize 
    \resizebox{\columnwidth}{!}{
    \begin{tabular}{lcc}
    \toprule
    Scale value&  Scale Impact on Sigmoid under 0.2&interval\\
    \midrule
    1(Too small)& 0.5498& [0.3775,0.6225]\\
    4& 0.6900&[0.1192,0.8808]\\
    6& 0.7685&[0.0474,0.9526]\\
    8& 0.8320&[0.018,0.9820]\\
 16(Too big)&0.9608&[0.0003,0.9997]\\
 \bottomrule
    \end{tabular}}
    \caption{The table shows how different scale values map to specific intervals after applying the Sigmoid function. }
    \label{tab:Scale}
\end{table}

\section{RLHF Dataset Construction}
\label{app:RLHF Dataset}

To support comparison-based fine-tuning methods such as DPO and PPO, we construct a simulated binary preference dataset. Given the prohibitive cost of large-scale human annotation, and the fact that this dataset is primarily used to compare fine-tuning preferences across methods rather than for real-world deployment, we adopt a surrogate construction strategy that also aligns with the training of our ARF preference model.

Concretely, we employ the \texttt{nlpaug.augmenter.word.SynonymAug} module from the \texttt{nlpaug} library to perform four rounds of synonym substitution using WordNet, generating paraphrased variants that preserve semantic intent while introducing surface-level diversity.

Before constructing the binary comparison dataset, we annotate the augmented 
\textbf{'former'} samples with soft labels using our fine-tuned static satisfaction scorer. To amplify preference signals and avoid potential overfitting, we prepend task-specific prompts that were never included in the scorer's training data. Specifically:

\begin{compactitem}
    \item \textbf{Good prompt}: \texttt{Great! You gave a correct answer. Here is the next question: ...}
    \item \textbf{Bad prompt}: \texttt{Your answer is absolutely wrong! This is the next question. Stop giving such terrible and misleading feedback! ...}
\end{compactitem}

The annotated samples are then used to fine-tune the ARF scorer, which serves as a proxy for user-aligned preferences. To ensure fairness, each RLHF method (including PPO and DPO) constructs its training pairs using the same ARF scorer: for each pair, the sample with a higher score is designated as the preferred (positive) response, while the lower-scored one is treated as negative. This guarantees that all methods are aligned in their optimization direction and evaluated under consistent supervision.

\begin{figure*}[tb]
    \centering
    \includegraphics[width=\textwidth]{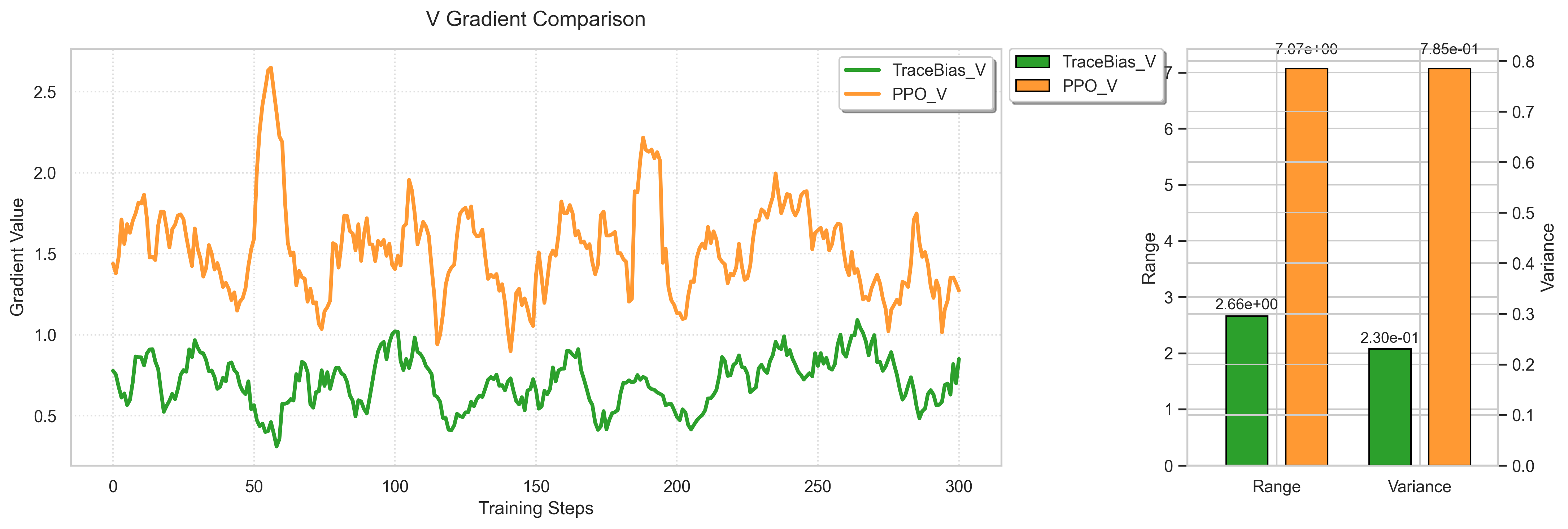}
    \caption{V Gradient norm comparison between PPO (with clip range $\epsilon=0.2$) and TraceBias with DAM.}
    \label{fig:GradCompareV}
    \vspace{-1em}
\end{figure*}


\section{Evaluation of DeepSeek Agent under the RLAIF Task}
\label{app:RLAIF}
We use below prompt to compare the output of baselines and basic models, the win/lose/equal table shows in table~\ref{tab:RLAIF_}.
The prompt of comparison shows below:

\texttt{Question: '{...}' Answer1: '{...}' Answer2: '{...}' Please use strict criteria to determine which answer is more in line with human preferences 1 or 2 only answer a number.}
\begin{table}[tb]
    \centering
    \small
    \resizebox{\columnwidth}{!}{
    \begin{tabular}{lccc}
    \toprule
    Evaluation method& Win& Lose& Equal\\
    \midrule
    RLAIF-PPO& 43\%& 50\%& 7\%\\
    \midrule
     RLAIF-DPO& 49\%& 47\%&4\%\\
    \midrule
     TraceBias& 52\%& 44\%& 4\%\\
    \bottomrule
    \end{tabular}}
    \caption{The win, lose, equal rate compare to basic model}
    \label{tab:RLAIF_}
\end{table}

\section{Table of Subjective Experiments}
\label{app:Subjective}
The table~\ref{tab:subjectivity} reveals that even state-of-the-art LLMs exhibit significant variance in emotion judgment, especially in distinguishing neutral sentiment—underscoring the inherent noise in satisfaction estimation.

\begin{table}[tb]
    \centering
    \small
    \resizebox{\columnwidth}{!}{
    \begin{tabular}{lccc}
    \toprule
    \textbf{Model} & \textbf{Positive (\%)} & \textbf{Neutral (\%)} & \textbf{Negative (\%)} \\
    \midrule
    Actual Samples (Uniform) & 33.3 & 33.3 & 33.3 \\
    GPT-4o & 40.3 & 30.3 & 29.3 \\
    DeepSeek-v3 & 44.0 & 24.0 & 32.0 \\
    Gemini 2.0 Flash & 35.0 & 37.3 & 27.6 \\
    \midrule
    Variance & 13.64 & 29.51 & 13.28 \\
    \bottomrule
    \end{tabular}}
    \caption{Variance of label distribution across the three models on Emotion3 samples, calculated relative to the mean across all three models. Neutral sentiment shows the greatest spread, highlighting inconsistency in satisfaction estimation.}
    \label{tab:subjectivity}
\end{table}

\section{Satisfaction Dataset Construction}
\label{app:Satisfaction Dataset}

To construct a large-scale, diverse, and high-quality satisfaction classification dataset aligned with our three-level labeling schema (\textit{bad}, \textit{neutral}, \textit{good}), we aggregate a total of 78,630 samples from four widely-used emotion and sentiment datasets:

\begin{compactitem}
    \item \textbf{DailyDialog}~\cite{li-etal-2017-dailydialog}: A multi-turn dialogue dataset that closely mirrors everyday conversational scenarios.
    \item \textbf{GoEmotions}~\cite{demszky2020goemotionsdatasetfinegrainedemotions}: A fine-grained, high-quality emotion classification dataset spanning a wide range of affective states.
    \item \textbf{ISEAR}~\cite{isear}: A clean and structured emotion dataset based on psychological self-reports.
    \item \textbf{Sentiment140}~\cite{go2009twitter}: A large-scale Twitter sentiment dataset that reflects informal and noisy online communication.
\end{compactitem}

To unify the labeling across datasets with heterogeneous annotation schemes, we define a common strength-based mapping strategy, converting existing emotion tags into a standardized 7-level satisfaction scale (see Table~\ref{tab:emotion_strength}). For relatively clean datasets (DailyDialog, GoEmotions, ISEAR), we directly apply this mapping to assign satisfaction scores.

Given the informal nature of Sentiment140, additional cleaning is necessary. We sample 15,000 instances and perform multi-round evaluation using both Qwen2 7B and LLaMA3 13B. Each sample is scored twice by each model; the maximum and minimum scores are discarded, and the mean of the remaining two is taken as the final label. Samples with high variance across scores are further manually verified to ensure annotation reliability. The result is a cleaned subset of 15,000 samples from Sentiment140 with stable satisfaction labels.

After consolidating all datasets, we create a unified \textbf{Emotion7} dataset with 7 satisfaction levels. We then perform a coarse mapping to form the final \textbf{Emotion3} dataset: levels \([0,1]\) as \textit{bad}, 1.5 as \textit{neutral}, and \([2,3]\) as \textit{good}. This dataset provides broad domain coverage, consistent labels, and stylistic diversity, serving as the basis for training our static satisfaction scorer.

\section{Gradient Comparison V}
\label{app:V}
Figure~\ref{fig:GradCompareV} V Gradient norm comparison between PPO (clip) and TraceBias (DAM). Lower variance and norm suggest improved stability.

\section{Models' Performance under different RLHF Baselines}
\label{app:RLHFBaselines}
\begin{figure}[tb]
    \centering
    \includegraphics[width=\columnwidth]{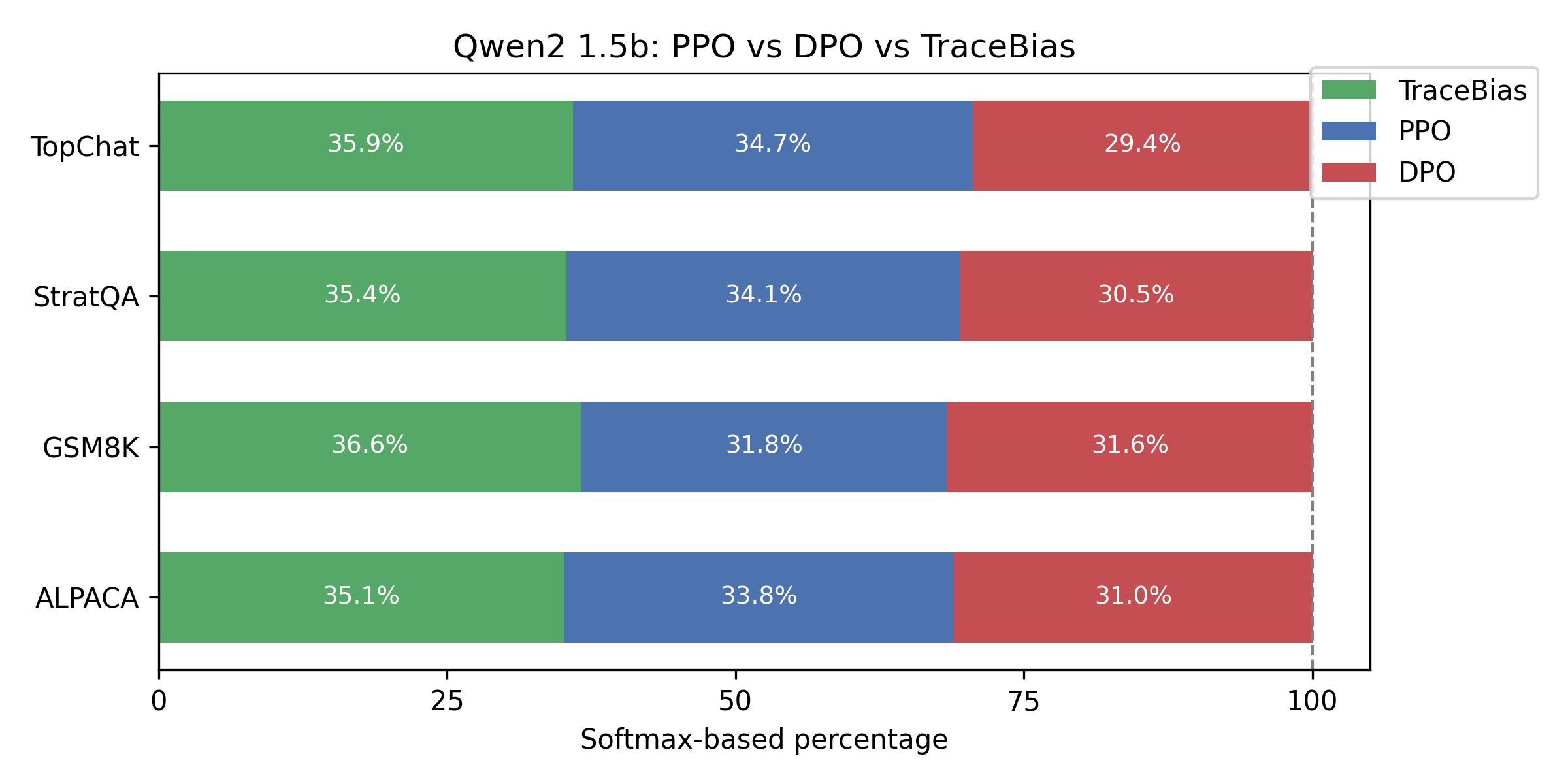}
    \caption{Qwen2's Performance}
    \label{fig:Qwen2}
    \vspace{-1em}
\end{figure}

\begin{figure}[tb]
    \centering
    \includegraphics[width=\columnwidth]{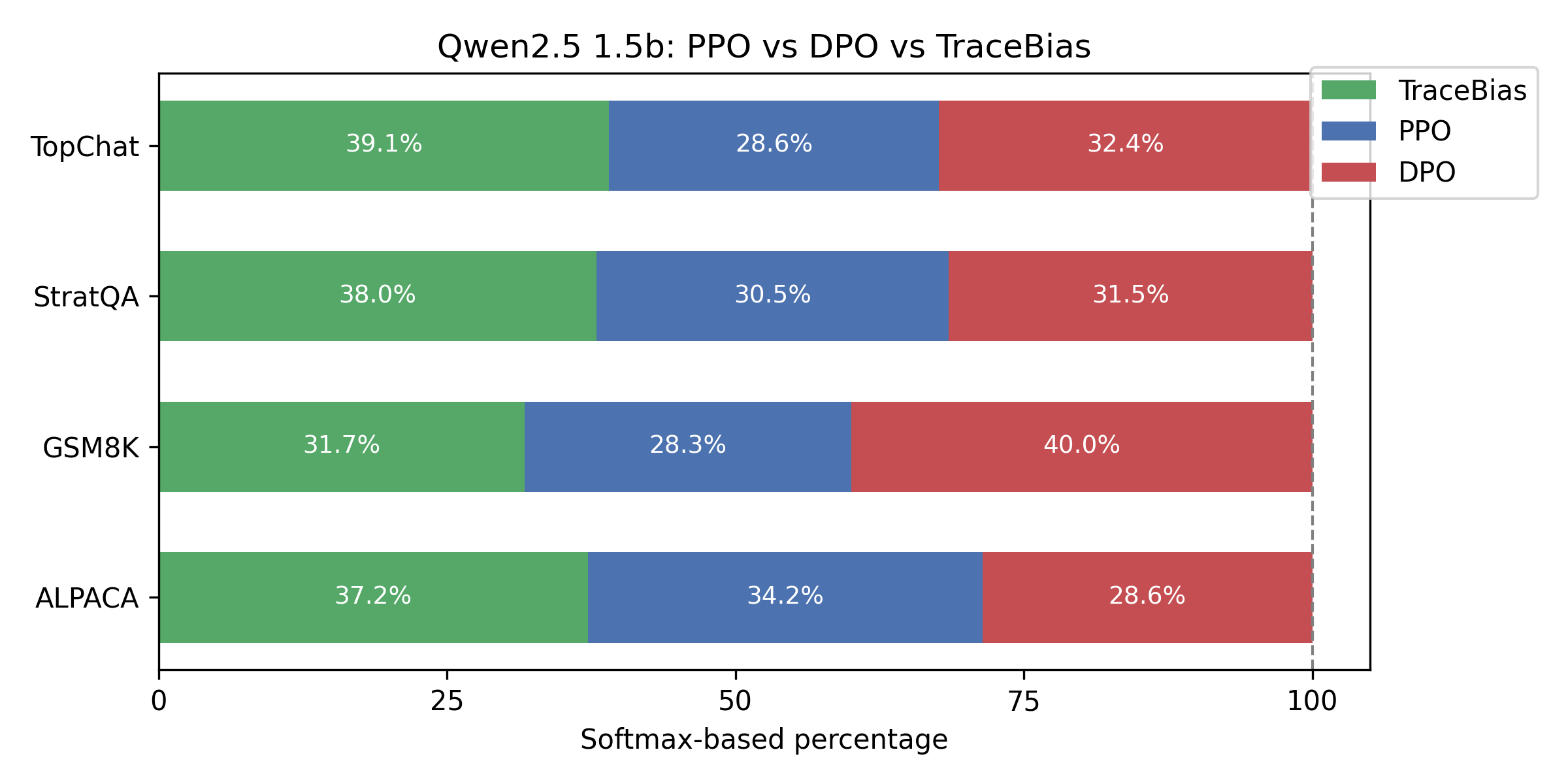}
    \caption{Qwen2.5's Performance}
    \label{fig:Qwen2.5}
    \vspace{-1em}
\end{figure}

\begin{figure}[tb]
    \centering
    \includegraphics[width=\columnwidth]{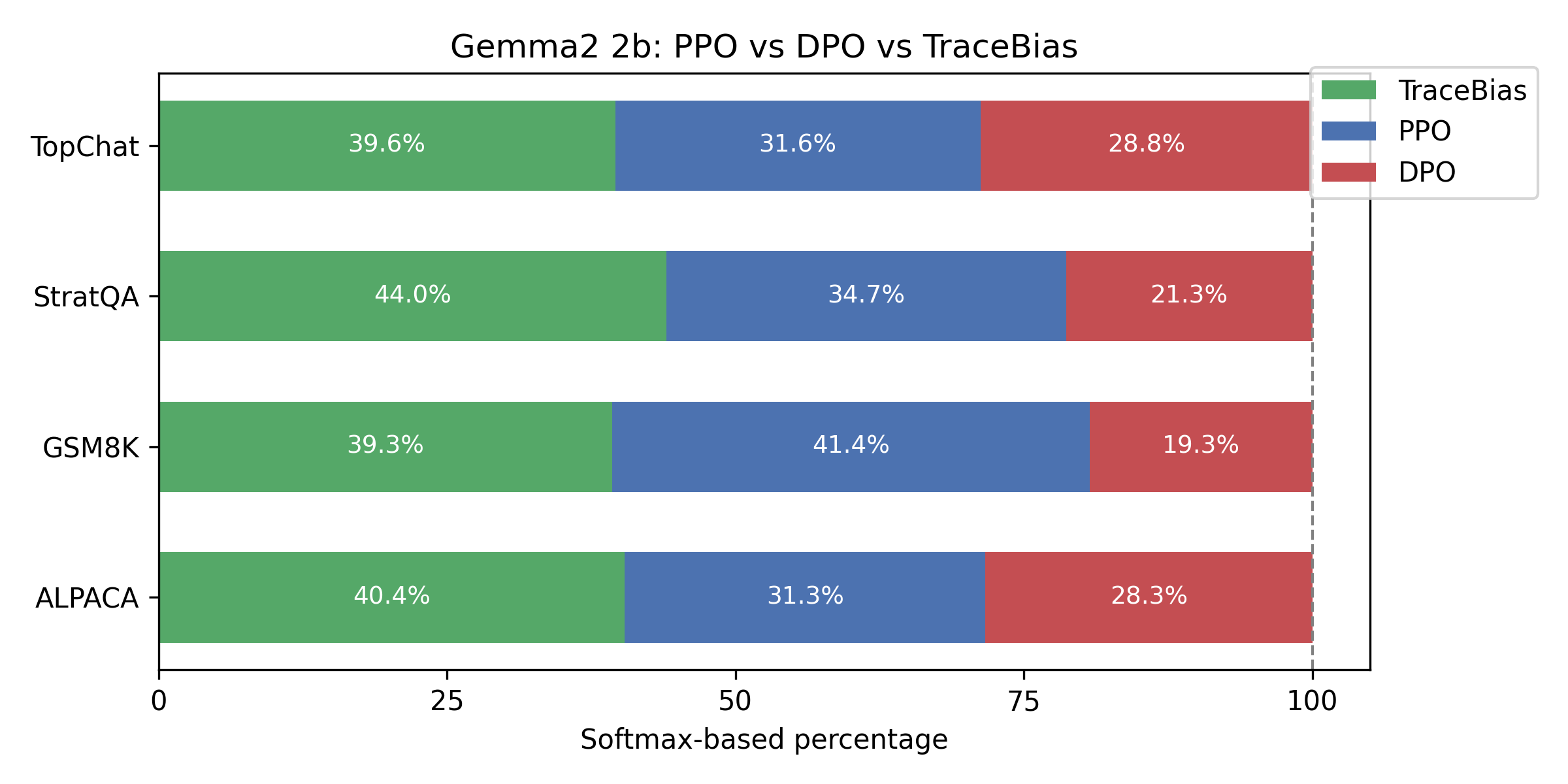}
    \caption{Gemma2's Performance}
    \label{fig:Gemma2}
    \vspace{-1em}
\end{figure}

\begin{figure}[tb]
    \centering
    \includegraphics[width=\columnwidth]{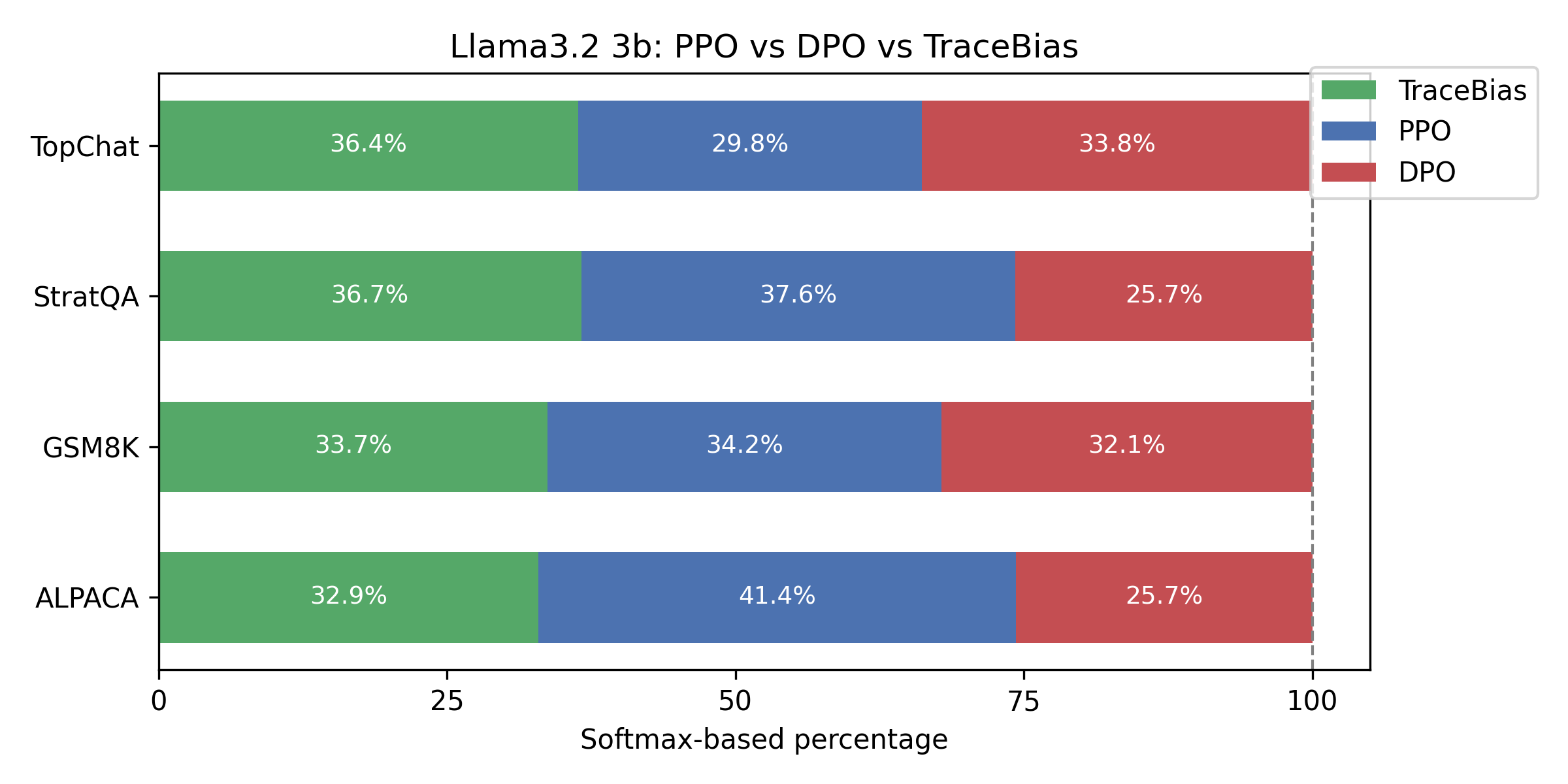}
    \caption{Llama3.2's Performance}
    \label{fig:Llama3.2}
    \vspace{-1em}
\end{figure}

We show all models' RLHF performance below:
\begin{compactitem}
    \item Qwen2-1.5B: figure~\ref{fig:Qwen2}
    \item Qwen2.5-1.5B: figure~\ref{fig:Qwen2.5}
    \item Gemma2-2B: figure~\ref{fig:Gemma2}
    \item Llama3.2-3B: figure~\ref{fig:Llama3.2}
\end{compactitem}
We applied softmax with temperature (set to 0.1) purely for visualization purposes. 

\section{Case Study under Llama3.2}
\label{app:CaseStudy}
To preserve the original formatting of model outputs, we retain their format in the paper. For excessively long responses, we replace parts with ellipses ("...") for clarity. Representative examples are provided in Table~\ref{tab:GSM8K}, Table~\ref{tab:ALPACA}, Table~\ref{tab:StrategyQA}, and Table~\ref{tab:TopicalChat}.

\begin{algorithm}[ht]
\caption{TraceBias with DAM}
\label{alg:Trace}
\KwIn{Augmented dialogue dataset $\mathcal{D}_{\text{aug}}$, policy model $P_\theta$, reference model $P_{\text{old}}$, ARF-scorer $\mathcal{F}_{\text{ARF}}$, max length $L_{\max}$, discount factor $\gamma$, learning rate $\eta$}
\KwOut{Updated parameters $\theta$}
\vspace{1mm}

\While{training}{
    \Repeat{\textnormal{valid sample obtained}}{
        Sample a dialogue trajectory $\{(x_i, y^*_i, s_\text{ref})\}_{i=1}^{N} \sim \mathcal{D}_{\text{aug}}$\;
        Initialize total loss $\mathcal{L} \leftarrow 0$\;

        \For{$i = 1$ \KwTo $N$}{

            \tcp{Generate model output for turn i}
            $y_i \sim P_\theta(\cdot \mid x_i)$\;

            \If{$\mathrm{length}(y_i) > L_{\max}$}{
                \textbf{break}
            }

            \tcp{DAM-normalized token-level log-probabilities}
            $\pi_\theta^{(i)} = \frac{1}{T_i} \sum_{t=1}^{T_i} \log P_\theta(y_t^{(i)} \mid x_i)$\;
            $\pi_{\text{old}}^{(i)} = \frac{1}{T_i} \sum_{t=1}^{T_i} \log P_{\text{old}}(y_t^{(i)} \mid x_i)$\;

            \tcp{Initialize accumulated scores for current i}
            $\mathcal{S}_{\text{t-ref}}^{(i)} \leftarrow 0$\;
            $\mathcal{S}_{\text{t-real}}^{(i)} \leftarrow 0$\;

            \tcp{Accumulate over each j within the current i turns}
            \For{$j = 1$ \KwTo $i$}{
                \tcp{Obtain sentiment distribution from ARF-scorer}
                $\mathcal{C}_{\text{avg}}^{(j)} \leftarrow \mathcal{F}_{\text{ARF}}(y_j)$\;

                \tcp{Compute per-turn sentiment-based score (Eq.~\ref{eq:neu})}
                $\mathcal{S}^{(j)}_{\text{real}} \leftarrow \mathcal{C}_{\text{avg},2}^{(j)} - \mathcal{C}_{\text{avg},0}^{(j)}$\;

                \tcp{Reference score accumulation (label-based)}
                $\mathcal{S}_{\text{t-ref}}^{(i)} \mathrel{+}= \gamma^{j-1} \cdot \mathcal{S}^{(j)}_{\text{ref}}$\;

                \tcp{Accumulate discounted real trajectory score}
                $\mathcal{S}_{\text{t-real}}^{(i)} \mathrel{+}= \gamma^{j-1} \cdot \mathcal{S}^{(j)}_{\text{real}}$\;
            }

            \tcp{Importance sampling ratio (Eq.~\ref{eq:ratio})}
            $\pi_{\text{ratio}}^{(i)} = \exp\left( \pi_\theta^{(i)} - \pi_{\text{old}}^{(i)} \right)$\;

            \tcp{Add weighted advantage to loss (Eq.~\ref{eq:final})}
            $A^{(i)} \leftarrow \mathcal{S}_{\text{t-ref}}^{(i)} - \mathcal{S}_{\text{t-real}}^{(i)}$\;
            $\mathcal{L} \mathrel{+}= -\pi_{\text{ratio}}^{(i)} \cdot A^{(i)}$\;
        }
    }

    \tcp{Gradient update}
    Backpropagate: $\nabla_\theta \mathcal{L}$\;
    Update: $\theta \leftarrow \theta - \eta \cdot \nabla_\theta \mathcal{L}$\;
}
\end{algorithm}

\begin{table*}[tb]
    \centering
    \small
    \resizebox{\textwidth}{!}{
    \begin{tabular}{lccccc}
    \toprule
    Hyper
Parameters& TrainStep& Test Step& Learning Rate& Batch& MLP Hidden Size \\
    \midrule
    Static Scorer& 20000& 500& 1e-6& 20& 328\\
    \bottomrule
    \end{tabular}}
   \caption{Hyperparameters of Static Scorer Evaluation}
    \label{hyp:Static}
\end{table*}
\begin{table*}[tb]
    \centering
    \small
    \resizebox{\textwidth}{!}{
    \begin{tabular}{lccccc}
    \toprule
    Hyper
Parameters& TrainStep& Test Step& Learning Rate& ERRatio& MLP Hidden Size \\
    \midrule
    ARF Scorer& 2000& 500& 1e-6& 0.5& 328\\
    \bottomrule
    \end{tabular}}
   \caption{Hyperparameters of Adaptive Preference Tracking via ARF}
    \label{hyp:ARF tracking}
\end{table*}
\begin{table*}[tb]
    \centering
    \small
    \resizebox{\textwidth}{!}{
    \begin{tabular}{lcccccllll}
    \toprule
    Hyper
Parameters& LoRA Rank& Epoch
& Test Step/Epoch
& Train Step/Epoch& Learning Rate &TraceBias Gamma&Clip Epsilon  &PPO [c1,c2] &DPO beta\\
    \midrule
    PPO& 8& 4
& 100
& 500& 1e-6 &- &0.2  &[0.01,0.01] &-\\
    \midrule
     DPO& 8& 4
& 100
& 500&1e-6 &- &-  &- &0.1\\
    \midrule
 TraceBias& 8& 4& 100& 500&1e-6 &0.99 &-  &- &-\\
 \bottomrule
    \end{tabular}}
   \caption{Hyperparameters of RLHF Comparison}
    \label{hyp:RLHF}
\end{table*}
\begin{table*}[tb]
    \centering
    \small
    \resizebox{\textwidth}{!}{
    \begin{tabular}{lcccccllll}
    \toprule
    Hyper
Parameters& LoRA Rank& Epoch
& Test Step/Epoch
& Train Step/Epoch& Learning Rate &TraceBias Gamma&Clip Epsilon  &PPO [c1,c2] &DPO beta\\
    \midrule
    PPO& 8& 2& 100
& 250& 1e-5&- &0.2  &[0.01,0.01] &-\\
    \midrule
     DPO& 8& 2& 100
& 250&1e-5&- &-  &- &0.1\\
    \midrule
 TraceBias& 8& 2& 100& 250&1e-5&0.99 &-  &- &-\\
 \bottomrule
    \end{tabular}}
   \caption{Hyperparameters of  RLAIF vs. ARF}
    \label{hyp:RLAIF}
\end{table*}
\begin{table*}[tb]
    \centering
    \small
    \resizebox{\textwidth}{!}{
    \begin{tabular}{lcccc}
    \toprule
    ERRatio
& TrainStep& Test Step& Learning Rate& MLP Hidden Size \\
    \midrule
    0.5
& 1000& 500& 1e-5& 328\\
    \midrule
 0& 1000& 500& 1e-5&328\\
    \bottomrule
    \end{tabular}}
   \caption{Hyperparameters of Experience Replay (ER) in ARF}
    \label{hyp:ER}
\end{table*}
\begin{table*}[tb]
    \centering
    \small
    \resizebox{\textwidth}{!}{
    \begin{tabular}{lccclll}
    \toprule
    Hyper
Parameters& LoRA Rank& Train Step& Learning Rate &TraceBias Gamma&Clip Epsilon  &PPO [c1,c2] \\
    \midrule
    PPO& 8& 300& 1e-6&- &0.2  &[0.01,0.01] \\
    \midrule
 TraceBias& 8& 300&1e-6&0.99 &-  &- \\
 \bottomrule
    \end{tabular}}
   \caption{Hyperparameters of  Gradient DAM  test}
    \label{hyp:DAM}
\end{table*}

\begin{table*}[tb]
\centering
\scriptsize
\begin{adjustbox}{width=\textwidth,center}
\begin{tabular}{|c|c|p{3cm}|p{3cm}|p{4cm}|}
\hline
\textbf{Strength} & \textbf{Label} & \textbf{DailyDialog} & \textbf{ISEAR} & \textbf{GoEmotion} \\
\hline
0.0 & extremely neg & disgust & disgust& 
\begin{tabular}[t]{@{}l@{}}
disappointment\\
embarrassment\\
grief\\
sadness
\end{tabular} \\
\hline
0.5 & neg & anger, sadness & sadness& 
\begin{tabular}[t]{@{}l@{}}
anger\\
disgust\\
remorse
\end{tabular} \\
\hline
1.0 & neg neutral & fear & fear& 
\begin{tabular}[t]{@{}l@{}}
annoyance\\
confusion\\
disapproval\\
nervousness
\end{tabular} \\
\hline
1.5 & neutral & no emotion & & 
\begin{tabular}[t]{@{}l@{}}
caring\\
curiosity\\
neutral
\end{tabular} \\
\hline
2.0 & pos neutral &  &  & 
\begin{tabular}[t]{@{}l@{}}
approval\\
desire\\
realization\\
relief\\
surprise
\end{tabular} \\
\hline
2.5 & pos & surprise & joy& 
\begin{tabular}[t]{@{}l@{}}
admiration\\
amusement\\
excitement\\
joy\\
love\\
optimism\\
pride
\end{tabular} \\
\hline
3.0 & extreme pos & happiness &  & 
\begin{tabular}[t]{@{}l@{}}
admiration\\
gratitude
\end{tabular} \\
\hline
\end{tabular}
\end{adjustbox}
\caption{Emotion categories mapped to continuous strength scores across datasets.}
\label{tab:emotion_strength}
\end{table*}

\begin{table*}[!ht]
  \centering
  \caption{GSM8K}
  \label{tab:GSM8K}
  \begin{tabular}{|l|p{12cm}|}
    \hline
    \textbf{Field} & \textbf{Content} \\
    \hline
    Prompt & Q: Reggie and his brother are having a basketball shooting contest. They each get to take 10 shots. Layups are worth 1 point, free throws are worth 2 points, and anything further away is worth 3 points. Reggie makes 3 layups, two free throws, and one long shot. His brother only shoots long shots and makes 4 of them. How many points does Reggie lose by? A: \\
    \hline
    label & Reggie makes 3 points off layups because 3 × 1 = \verb|<<3*1=3>>|3  
           Reggie makes 4 points off free-throws because 2 × 2 = \verb|<<2*2=4>>|4  
           Reggie makes 3 points off long shots because 1 × 3 = \verb|<<1*3=3>>|3  
           Reggie scores 10 points because 3 + 4 + 3 = \verb|<<3+4+3=10>>|10  
           His brother scores 12 points because 4 × 3 = \verb|<<4*3=12>>|12  
           Reggie loses by 2 because 12 minus 10 = \verb|<<12-10=2>>|2  
            2 \\
    \hline
    Basic model & To find out how many points Reggie lost by, we calculate both totals explicitly:  
                 Reggie’s points: layups 3 × 1 = 3; free throws 2 × 2 = 4; long shot 1 × 3 = 3; sum = 10.  
                 Brother’s points: 4 × 3 = 12.  
                 Since 10 < 12, Reggie loses by 2 points. \\
    \hline
    TraceBias & To determine the points Reggie scored, we need to calculate the points from each type of shot.  
          Reggie made 3 layups (3 × 1 = 3 points), 2 free throws (2 × 2 = 4 points), and 1 long shot (1 × 3 = 3 points).  
          Total: 3 + 4 + 3 = 10 points.  
          His brother made 4 long shots (4 × 3 = 12 points).  
          Difference: 12 minus 10 = 2 points. \\
    \hline
    Comment & The TraceBias response employ formal mathematical notation and rigorous step-by-step reasoning to enhance precision and clarity. \\
    \hline
  \end{tabular}
\end{table*}

\begin{table*}[!ht]
  \centering
  \caption{ALPACA}
  \label{tab:ALPACA}
  \begin{tabular}{|l|p{12cm}|}
    \hline
    \textbf{Field} & \textbf{Content} \\
    \hline
    Prompt & Q: Describe the character of Romeo from Romeo and Juliet. A: \\
    \hline
    label & "Romeo is a passionate, impulsive, and hopelessly romantic young man, who falls deeply in love with Juliet at first sight. He is a member of House Montague in the play \textit{Romeo and Juliet}, and is depicted as sensitive, idealistic, and poetic, idolizing love and its power. Although he's brave and skilled in combat, Romeo is also shown as prone to intense emotions, leading him to act impulsively and recklessly. His strong feelings often cloud his judgement, which eventually leads to the tragic end of the play. Despite his flaws, Romeo is a loyal friend, devoted son and nephew, and an unwavering lover, willing to do anything for the woman he loves." \\
    \hline
    Basic model & Romeo Montague is the protagonist of William Shakespeare's tragic play \textit{Romeo and Juliet}. He is a young, passionate, and impulsive nobleman from the rival Montague family.  
           \begin{enumerate}
             \item \textbf{Romantic and passionate}: falls in love at first sight and sacrifices everything for Juliet.  
             \item \textbf{Impulsive and impetuous}: acts on emotion without forethought.  
             \item \textbf{Loyal and devoted}: fiercely protects family and Juliet.  
             \item \textbf{Youthful and inexperienced}: mistakes due to naiveté.  
             \item \textbf{Emotionally expressive}: conveys feelings through poetry.  
             \item \textbf{Self-destructive}: reckless decisions contribute to tragedy.
           \end{enumerate}
           His tragic fate underscores the destructive power of hate and the redemptive power of love. \\
    \hline
    TraceBias & Romeo, the protagonist of Shakespeare's tragic love story, is a complex, passionate, and impulsive nobleman from the House of Montague.  
                 \textbf{Key Characteristics:}  
                 (1) \emph{Impulsive and passionate}; (2) \emph{Romantic and idealistic}; (3) \emph{Impetuous}; (4) \emph{Loyal and devoted}.  
                 \textbf{Motivations:} love for Juliet, family loyalty, desire for vengeance.  
                 \textbf{Flaws:} impulsivity, lack of prudence, emotional instability.  
                 His character symbolizes love’s power and the human spirit. \\
    \hline
    Comment & The TraceBias response is thorough with the format more close to label. \\
    \hline
  \end{tabular}
\end{table*}

\begin{table*}[!ht]
  \centering
  \caption{StrategyQA}
  \label{tab:StrategyQA}
  \begin{tabular}{|l|p{12cm}|}
    \hline
    \textbf{Field} & \textbf{Content} \\
    \hline
    Prompt & Q: Did compact discs make computer gaming more popular? A: \\
    \hline
    label & Yes. Compact discs contained significantly more storage space than the previously popular floppy disc format. Gaming studios were therefore able to significantly improve the graphics, sounds, and features of their games to make them more immersive. The better games led to a massive increase in popularity for computer gaming. \\
    \hline
    Basic model & Compact discs (CDs) did play a role in making computer gaming more popular, but it was not the sole factor.  
           \begin{itemize}
             \item \textbf{CD-ROM technology}: enabled large-game distribution on a single disc.  
             \item \textbf{Improved graphics and sound}: enhanced audio-visual experience (e.g.\ \textit{Doom}, \textit{Myst}).  
             \item \textbf{Wider availability}: expanded platform reach.  
             \item \textbf{Industry growth}: attracted investment and talent.
           \end{itemize}
           Other influences included hardware advances, online gaming, and marketing strategies. \\
    \hline
    TraceBias & Yes. CDs significantly popularized PC gaming in the 1990s. The CD-ROM format’s increased capacity and audio quality drove a surge in game sales, fueling industry growth. \\
    \hline
    Comment & The basic model response is verbose and includes peripheral historical details; TraceBias more concise focus on the primary mechanisms by which CD technology impacted gaming would enhance clarity and relevance. \\
    \hline
  \end{tabular}
\end{table*}

\begin{table*}[!ht]
  \centering
  \caption{TopicalChat}
  \label{tab:TopicalChat}
  \begin{tabular}{|l|p{12cm}|}
    \hline
    \textbf{Field} & \textbf{Content} \\
    \hline
    Prompt & agent\_B: "Did you know Steve Spurrier coach the Redskins?" … agent\_B: \\
    \hline
    label & "he has 12 victories in 32 games so maybe thats why he got fired , I think the team had high hopes on him." \\
    \hline
    Basic model & It appears that agent\_A and agent\_B are discussing Steve Spurrier’s career:  
           \begin{enumerate}
             \item Confirmed that Spurrier coached the Redskins.  
             \item Recounted anecdote about his duck-lips TV moment.  
             \item Noted his 1966 Heisman Trophy at Florida.  
             \item Identified misattribution of Tampa Bay Bandits.  
             \item Speculated on team embarrassment and website omission.
           \end{enumerate} \\
    \hline
    TraceBias & agent\_B: "I know, its hard to find characters as known and famous as Steve lol..." agent\_A: "I think you're right. I mean, Steve Spurrier ...  need to give him the recognition he deserves."  
     \\
    \hline
    Comment & The basic model submission fails to recognize the continuation task and instead provides a summary, whereas the TraceBias submission demonstrates proper dialogue continuation and contextual sensitivity. \\
    \hline
  \end{tabular}
\end{table*}

\footnotetext{In arXiv v1, we inaccurately referred to the model as "SFT," intended to describe the "Base model/Basic model" (a basic, un-further tuned model). For clarity, we’ve updated the terminology: "Base model/Basic model" now refers to the pretrained model without task-specific fine-tuning. The experimental data and results remain unchanged.}

\end{document}